\pdfoutput=1
\documentclass[11pt]{article}
\usepackage{graphicx}
\usepackage{subcaption}
\usepackage[final]{acl}

\usepackage{times}
\usepackage{latexsym}
\usepackage{amsmath}
\usepackage{verbatim} 
\usepackage{makecell}
\definecolor{lightblue}{RGB}{135, 206, 250} % 稍深的蓝色
\definecolor{lightorange}{RGB}{255, 165, 79} % 深化橙色
\definecolor{lightgreen}{RGB}{102, 205, 102} % 更深的绿色
\definecolor{lavender}{RGB}{200, 160, 230} % 稍深的紫色
\definecolor{lightyellow}{RGB}{255, 215, 102} % 深化黄色
\usepackage[T1]{fontenc}
\usepackage[utf8]{inputenc}
\usepackage{enumitem}
\usepackage{microtype}
\usepackage{tcolorbox}
\usepackage{multirow}
\usepackage{inconsolata}
\usepackage{tikz}

\usepackage{amsfonts}
\usepackage{amsmath,amssymb,bm}
\usepackage{float}
\usepackage{xcolor}
\definecolor{darkgreen}{RGB}{0,100,0} % 深绿色
\usepackage{mdframed}
\usepackage{tcolorbox}
\newtcolorbox{response}[1]{colback=gray!10!white, colframe=black!80!white, title={#1}}

\usepackage{enumitem}
\usepackage{listings}
\newtcolorbox{promptbox}[1][]{
    colback=white,
    colframe=black!50,
    fonttitle=\bfseries,
    title=#1,
    left=6pt,
    right=6pt,
    top=8pt,
    bottom=8pt,
    arc=2pt
}

\usepackage{CJKutf8}
\usepackage{color}

\title{Cross-Document Cross-Lingual NLI via RST-Enhanced Graph Fusion and Interpretability Prediction}

\author{
  \textbf{Mengying Yuan\textsuperscript{1*}},
  \textbf{Wenhao Wang\textsuperscript{2*}},
  \textbf{Zixuan Wang\textsuperscript{1}},
  \textbf{Yujie Huang\textsuperscript{1}},
\\
  \textbf{Kangli Wei\textsuperscript{1}},
  \textbf{Fei Li\textsuperscript{1}\textsuperscript{\textdagger}},
  \textbf{Chong Teng\textsuperscript{1}},
  \textbf{Donghong Ji\textsuperscript{1}},
\\
  \textsuperscript{1}Key Laboratory of Aerospace Information Security and Trusted Computing,\\
  Ministry of Education, School of Cyber Science and Engineering, Wuhan University \\
  \textsuperscript{2}Zhejiang University
\\
  \small{\{yuanmengying\_51,zixuanwang\_nlp,huang-yj,kangliwei,lifei\_csnlp,tengchong,dhji\}@whu.edu.cn}
}%\author{
\usepackage{booktabs}
\begin{document}
\maketitle

\begingroup
\renewcommand\thefootnote{$*$}
\footnotetext{Equal contribution.}
\renewcommand\thefootnote{\textdagger}
\footnotetext{Corresponding author.}

\endgroup

\begin{abstract}
Natural Language Inference (NLI) is a fundamental task in natural language processing. While NLI has developed many subdirections such as sentence-level NLI, document-level NLI and cross-lingual NLI, Cross-Document Cross-Lingual NLI (CDCL-NLI) remains largely unexplored.
In this paper, we propose a novel paradigm: CDCL-NLI, which extends traditional NLI capabilities to multi-document, multilingual scenarios. 
To support this task, we construct a high-quality CDCL-NLI dataset including 25,410 instances and spanning 26 languages.
To address the limitations of previous methods on CDCL-NLI task, we  further propose an innovative method that integrates RST-enhanced graph fusion with interpretability-aware prediction.
Our approach leverages RST (Rhetorical Structure Theory) within heterogeneous graph neural networks for cross-document context modeling, and employs a structure-aware semantic alignment based on lexical chains for cross-lingual understanding. For NLI interpretability, we develop an EDU (Elementary Discourse Unit)-level attribution framework that produces extractive explanations.
Extensive experiments demonstrate our approach's superior performance, achieving significant improvements over both conventional NLI models
as well as large language models.
Our work sheds light on the study of NLI and will bring research interest on cross-document cross-lingual context understanding, hallucination elimination and interpretability inference.
Our code and dataset are available at \href{https://github.com/Leonardo123-ui/CDCL_NLI}{CDCL-NLI-link}.
\end{abstract}

\section{Introduction}

Natural Language Inference (NLI) is a fundamental task in natural language processing, aiming to determine the logical relationship between the given premise and hypothesis pair~\cite{dagan2005pascal, maccartney2009nli}.
While traditional NLI tasks primarily deal with single-language, short-text validations~\cite{Satoshi2007UNED}, document-level NLI~\cite{yin2021docnli} expands the scope of NLI to longer contexts.
% , encompassing hundreds or even thousands of tokens within a single document. 

\begin{figure}[t]
    \centering
    % \caption{A CDCL-NLI example to show how to determine the Entailment label by jointly reasoning from two premise documents.} 
    \includegraphics[width=\linewidth]{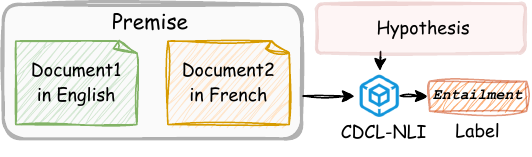}
    \vspace{-4mm}
    \caption{A CDCL-NLI example. Premise in \textcolor{darkgreen}{English} and \textcolor{orange}{French}. The Entailment label requires combining information from both documents in premise.}
    \label{fig:scenario}
\end{figure}
\begin{table}[t]
\small  
%  by premise, hypothesis, and language.
\setlength{\tabcolsep}{5pt}
\begin{tabular}{l|ccc}
\toprule
\textbf{Paradigm} & \textbf{Premise} & \textbf{Hypothesis} & \textbf{Language} \\
\midrule
Sentence-NLI & Sentence & Sentence & Mono/Multi \\
Document-NLI & Doc & Sent/Doc & Mono \\
CDCL-NLI & Multi Doc& Sentence & Multi \\
\bottomrule
\end{tabular}
\caption{Comparison of different NLI paradigms.}
\label{tab:nli-comparison}

\end{table}
Table~\ref{tab:nli-comparison} compares different NLI paradigms systematically,  highlighting the progressive evolution of NLI tasks. Sentence-NLI involves low-complexity reasoning on short sentence pairs, evolves from single-language approaches~\cite{bowman2015snli, herlihy2021mednli} to multilingual settings~\cite{conneau2018xnli,heredia2024xnlieu}, and is mainly used for fact verification~\cite{wadden2020fact,SI-NLI}. 
% , nie2020anli, hu2020xtreme thorne2018fever
% Document-level NLI expands to document-level reasoning, within a single document and a single language~\cite{Wang2019nliqa, yin2021docnli}, focusing on content comprehension~\cite{dua2019drop}.  
Document-level NLI extends NLI to reasoning over full-length documents within a single language~\cite{Wang2019nliqa, yin2021docnli}, focusing on content comprehension~\cite{yang2024supervised}. 

However, the increasing globalization of information flow requires even more sophisticated inference capabilities across both language and document boundaries. In this paper, we introduce \textbf{C}ross-\textbf{D}ocument \textbf{C}ross-\textbf{L}ingual \textbf{N}atural \textbf{L}anguage \textbf{I}nference (\textbf{CDCL-NLI}), a novel paradigm extending traditional NLI to multi-document and multilingual settings. Figure~\ref{fig:scenario} illustrates that CDCL-NLI jointly reasons over premise documents in English and French to verify the hypothesis. The correct Entailment prediction relies on integrating complementary information from both documents.
% As illustrated in Figure~\ref{fig:scenario}, CDCL-NLI jointly processes premise documents in Spanish and French to verify an English hypothesis. The Spanish document states that \textcolor{darkgreen}{Country X commits to reducing carbon emissions via renewable energy and UN cooperation}, while the French document adds that \textcolor{orange}{Country X will cooperate with neighboring countries to build a regional clean energy network}. The correct prediction of the label (Entailment) necessitates joint reasoning over both premise documents.

% Part of premise mentions \textcolor{darkgreen}{Country X’s commitment to reducing carbon emissions via renewable energy and UN cooperation}, while the other states \textcolor{orange}{cooperation with neighbors to build a clean energy network}.

%     \caption{A CDCL-NLI example to show how to determine the Entailment label by jointly reasoning from two premise documents. Dashed box translations are for illustration only and not used in reasoning.}
While CDCL-NLI addresses a real-world task with broad applications, it faces key challenges:
\textbf{1) Lack of existing datasets}, which necessitates the construction of new resources to support research.
\textbf{2) Multilingual Semantic Alignment}, requiring resolution of grammatical and conceptual differences across languages while preserving semantic consistency~\cite{Conneau2020uclr}.
 \textbf{3) Cross-Document Structure Alignment}, essential for capturing structural correspondences and implicit logical relations between documents of varying complexity~\cite{Wang2021satr}; and \textbf{4) Interpretability}, demanding transparent reasoning processes and verifiable confidence in inference outcomes~\cite{bereska2024mechanistic}.

% To address these challenges, firstly, we focus on building a high-quality dataset. \textbf{1) Dataset Construction:} To facilitate research on CDCL-NLI, we curated a multilingual dataset through three steps: collecting diverse premise documents from GlobeSumm~\cite{ye2024globesumm}; generating hypotheses with GPT-4o~\cite{openai2024gpt4} using customized prompts to ensure label diversity and balance; manually reviewing hypotheses and annotated explanations. The dataset contains 25,410 samples spanning 26 languages and 370 news events. 
% Secondly, we propose a novel method that comprises three designed key components which can address the rest challenges respectively.
% \textbf{2) Graph Construction Module:} This component promotes semantic alignment by fusing graphs based on lexical chains, effectively linking semantically related concepts across documents.
% \textbf{3) Graph Representation Module:} Utilizing an RST-enhanced Relation-aware Graph Attention Network (RGAT)~\cite{mann1988rhetorical, busbridge2019rgat}, this module supports structure alignment by capturing hierarchical discourse structures and cross-document dependencies through multi-head attention mechanisms.
% \textbf{4) Interpretability Attribution Module:} Leveraging Elementary Discourse Units (EDUs)~\cite{mann1988rhetorical}, this module generates extractive explanations that significantly enhance model interpretability and provide transparent insights into its decision-making process.

To address the first challenge, we curated a \textbf{CDCL-NLI dataset} through collecting diverse premise documents from GlobeSumm~\cite{ye2024globesumm}, generating hypotheses with GPT-4o~\cite{openai2024gpt4} using customized prompts to ensure label diversity and balance and manually reviewing hypotheses and annotated explanations. 
The dataset contains 25,410 samples spanning 26 languages and 370 events. 

To address the rest challenges, we proposed a novel method that comprises three key components.
\textbf{1) Graph Construction Module:} This component promotes semantic alignment by fusing graphs based on lexical chains, effectively linking semantically related concepts across documents.
\textbf{2) Graph Representation Module:} Utilizing an RST-enhanced Relation-aware Graph Attention Network (RGAT)~\cite{mann1988rhetorical, busbridge2019rgat}, this module supports structure alignment by capturing hierarchical discourse structures and cross-document dependencies through multi-head attention mechanisms.
\textbf{3) Interpretability Attribution Module:} Leveraging Elementary Discourse Units (EDUs)~\cite{mann1988rhetorical}, this module generates extractive explanations that significantly enhance model interpretability and provide transparent insights into its decision-making process.

Extensive experiments on the CDCL-NLI and DocNLI datasets demonstrate that our method outperforms conventional NLI approaches and three state-of-the-art large language models, surpassing the strongest baseline by 3.5\% on our dataset.
% Extensive experiments on CDCL-NLI dataset and DocNLI dataset demonstrate that our method outperforms conventinal NLI approaches such as DocNLI~\cite{yin2021docnli} and R2F~\cite{wang2022r2f}, as well as methods utilizing large language models, including Llama3-8B-Instruct~\cite{meta2024llama3}, Qwen-3~\cite{qwen3} and GPT-4o~\cite{openai2024gpt4}.
In the end, we highlight our main contributions as follows:

% \begin{itemize}[itemsep=2pt, topsep=4pt]
\begin{itemize}[nosep, left=0.5em]
    \item We propose CDCL-NLI as a new task and construct a corresponding dataset covering 26 languages with 25,410 high-quality manually-annotated instances.
    \item We propose a novel method that leverages RST-enhanced graph fusion to align semantic concepts and discourse structures. The approach also enhances interpretability by generating extractive, EDU-level explanations.
    \item We conduct extensive experiments demonstrating our method’s effectiveness, outperforming all baselines by at least 3.5\% and establishing a new benchmark for the CDCL-NLI task.
\end{itemize}

\section{Related Work}

\subsection{Sentence-level NLI}

\paragraph{Monolingual Methods.} 
Sentence-level NLI benchmarks like SNLI~\cite{bowman2015snli} and MultiNLI~\cite{williams2018multinli} have driven model evolution from ESIM~\cite{chen2017ESIM} to transformer architectures~\cite{devlin2018bert,liu2019roberta} and recent LLMs~\cite{openai2023gpt4}.

\paragraph{Cross-lingual Methods.}
Cross-lingual NLI relies on datasets like XNLI~\cite{conneau2018xnli} (15 languages) and XNLIeu~\cite{heredia2024xnlieu} (European languages). Multilingual models such as XLM-R~\cite{Conneau2020uclr} and XLM-E~\cite{chi2021xlm-e} enable zero-shot transfer, while alignment methods like SoftMV~\cite{hu2023softmv} and prompt-based MPT~\cite{qiu2024mpt} improve cross-lingual semantic understanding.

\paragraph{Interpretability Mechanisms.}
Interpretability uses feature attribution methods like Integrated Gradients~\cite{sundararajan2017axiomatic} and \cite{huang2024explainable} to highlight decision-driving features. Datasets such as e-SNLI~\cite{camburu2018e-snli} provide human explanations, supporting explicit reasoning and interpretability benchmarks.
\subsection{Document-level NLI}

\paragraph{Datasets and Benchmarks.}
Document-level NLI benefits from datasets like DocNLI~\cite{yin2021docnli} with over one million instances. Domain-specific datasets such as ContractNLI~\cite{koreeda2021contractnli} focus on the challenges posed by long documents and specialized text genres.

\paragraph{Inference Methods.}
Recent approaches emphasize discourse structure and long-range dependencies \cite{chen2025disretrieval}. R2F~\cite{wang2022r2f} introduces explicit reasoning extraction, and DocInfer~\cite{mathur2022docinfer} uses hierarchical encoding to model document structure, highlighting the need to capture document-level semantics.

\paragraph{Interpretability Mechanisms.}
Interpretability research focuses on evidence extraction and explanation generation. Systems like Evidence-Net~\cite{chen2022evidencenet} and R2F~\cite{wang2022r2f} automatically identify evidence to enhance reasoning transparency. LLM-based approaches like Chain-of-Thought~\cite{wei2022chain} and Rethinking \cite{singh2024rethinking} further enable self-explanatory reasoning capabilities.

\subsection{Graph-based Reasoning for NLI}
Leveraging graph structures for semantic reasoning has emerged as a powerful paradigm. Discourse-aware graph networks model logical relationships within text for tasks like logical reasoning~\cite{hou2022discourse, galitsky2025dialogue}. Similarly, AMR-based graph reasoning uses Abstract Meaning Representation (AMR) to enhance question answering by providing a structured semantic representation~\cite{huang2025amr}. Furthermore, prior work on graph merging and fusion has explored combining structures like AMR, RST, and CST for tasks such as multi-document summarization and inference~\cite{banarescu2018amr, liao2021docgraph, shi2024glimmer}.

Although prior studies have advanced sentence-level and document-level NLI, and graph-based methods have been applied to various reasoning tasks, the challenges in cross-document and cross-lingual NLI remain largely unaddressed. Our work fills this gap by introducing the CDCL-NLI dataset and proposing a systematic integration of an interpretable RST-enhanced graph fusion method to tackle these unique complexities.

\begin{figure}[t]
    \centering
    \includegraphics[width=\linewidth]{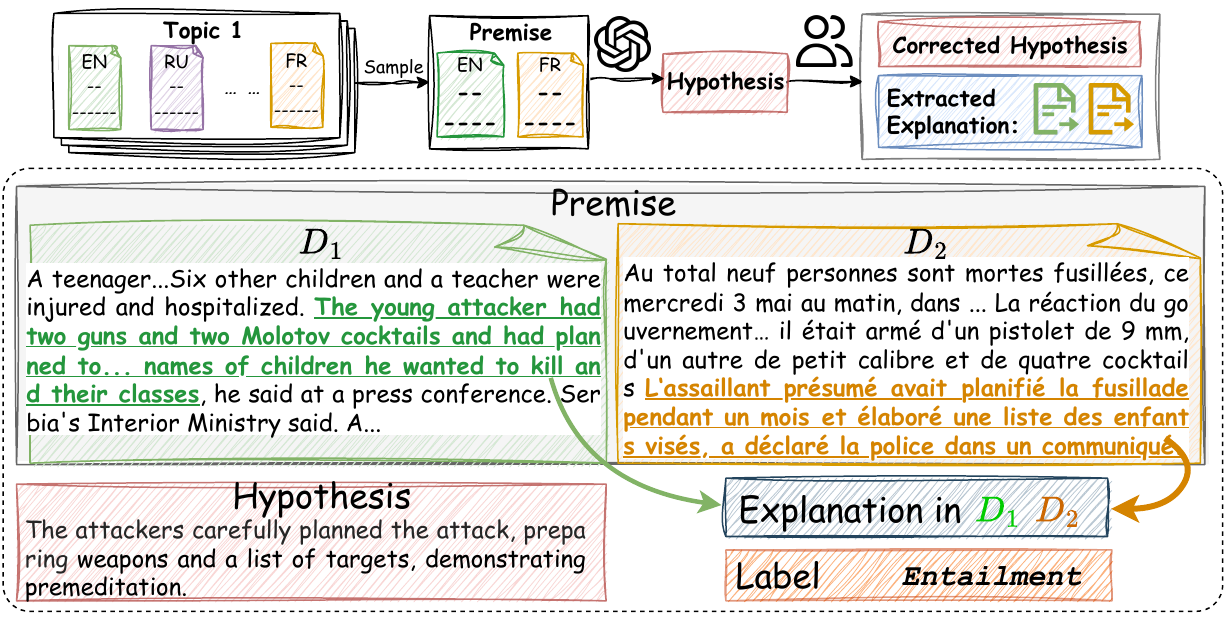}
    \vspace{-3mm}
    \caption{Overview of the CDCL-NLI dataset construction process and a data example.
    Premise contains \(D_1\) and  \(D_2\). Explanation is extracted from premise to enhance interpretability. Human annotation is based on language translated into English.}
    \label{fig:data}

\end{figure}

\section{CDCL-NLI Task Formulation and Dataset Construction}

\label{sec:data_construction}
As shown in Figure~\ref{fig:data}, our CDCL-NLI dataset is constructed through a systematic pipeline involving stratified random sampling of premise documents across all topics, LLM-generated hypotheses, and human verification to ensure data quality. In the dashed box, the figure shows a CDCL-NLI instance with a premise of two documents in different languages, an English hypothesis, a label, and EDU-based explanations for interpretability.
\subsection{Task Formulation}
Similar to the traditional NLI task, the goal of CDCL-NLI is to determine the inference label: 

\begin{displaymath}
\small
\textit{Label} \in \{\textit{"Entailment"}, \textit{"Neutral"}, \textit{"Contradiction"}\}, 
\end{displaymath}

between a given premise $ \textit{P} $ and hypothesis $ \textit{H} $. Specifically, the premise $ \textit{P} $ consists of two documents $ D_1 $ and $ D_2 $, written in different languages but discussing the same topic.
% \begin{displaymath}
% \textit{P} = \{D_1, D_2\}.
% \end{displaymath}
\begin{figure*}[t]
   
    \centering
    \begin{subfigure}[b]{0.31\textwidth}
        \centering
        \includegraphics[width=\textwidth]{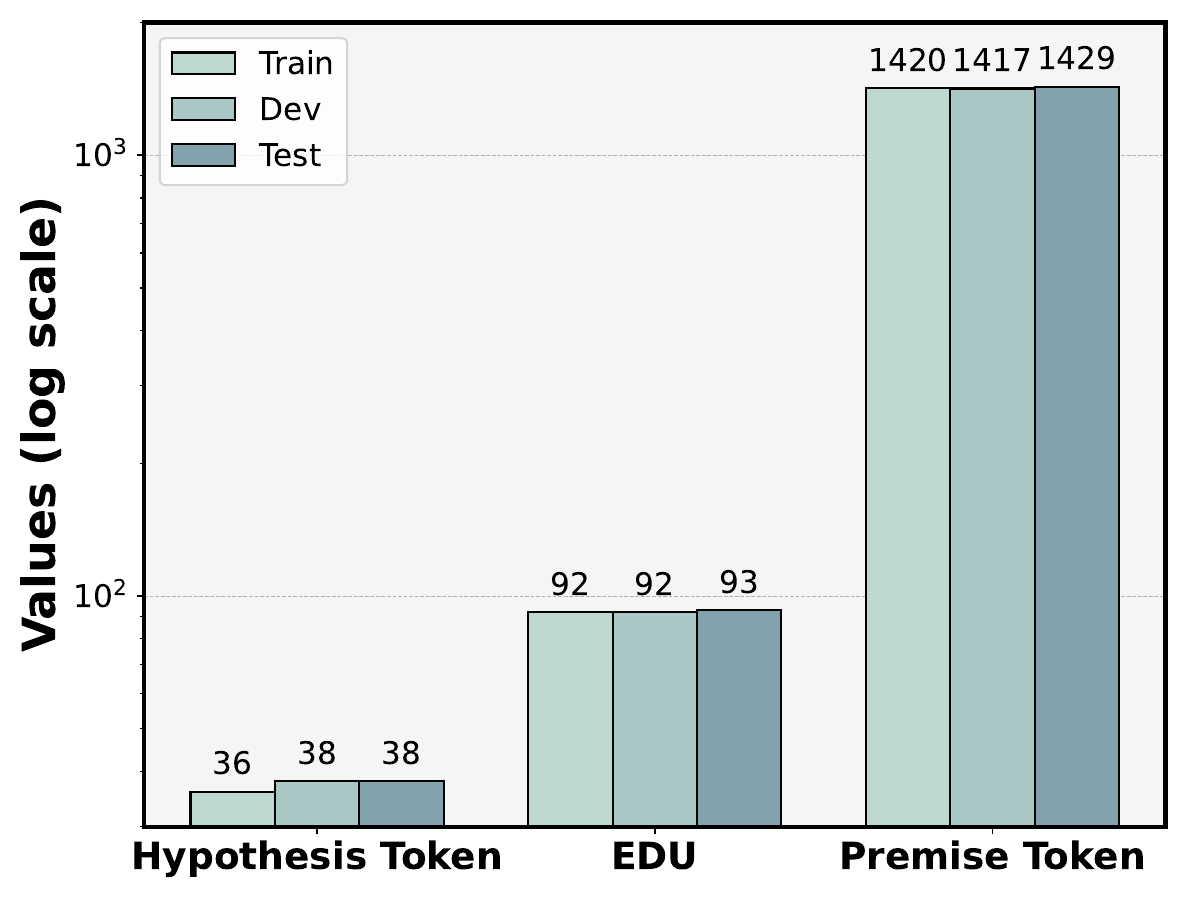}
        \caption{Log-scaled token count and EDU numbers across different data splits.}
        \label{fig:data_split}
    \end{subfigure}
    \hfill
    \begin{subfigure}[b]{0.39\textwidth}
        \centering
        \includegraphics[width=\textwidth]{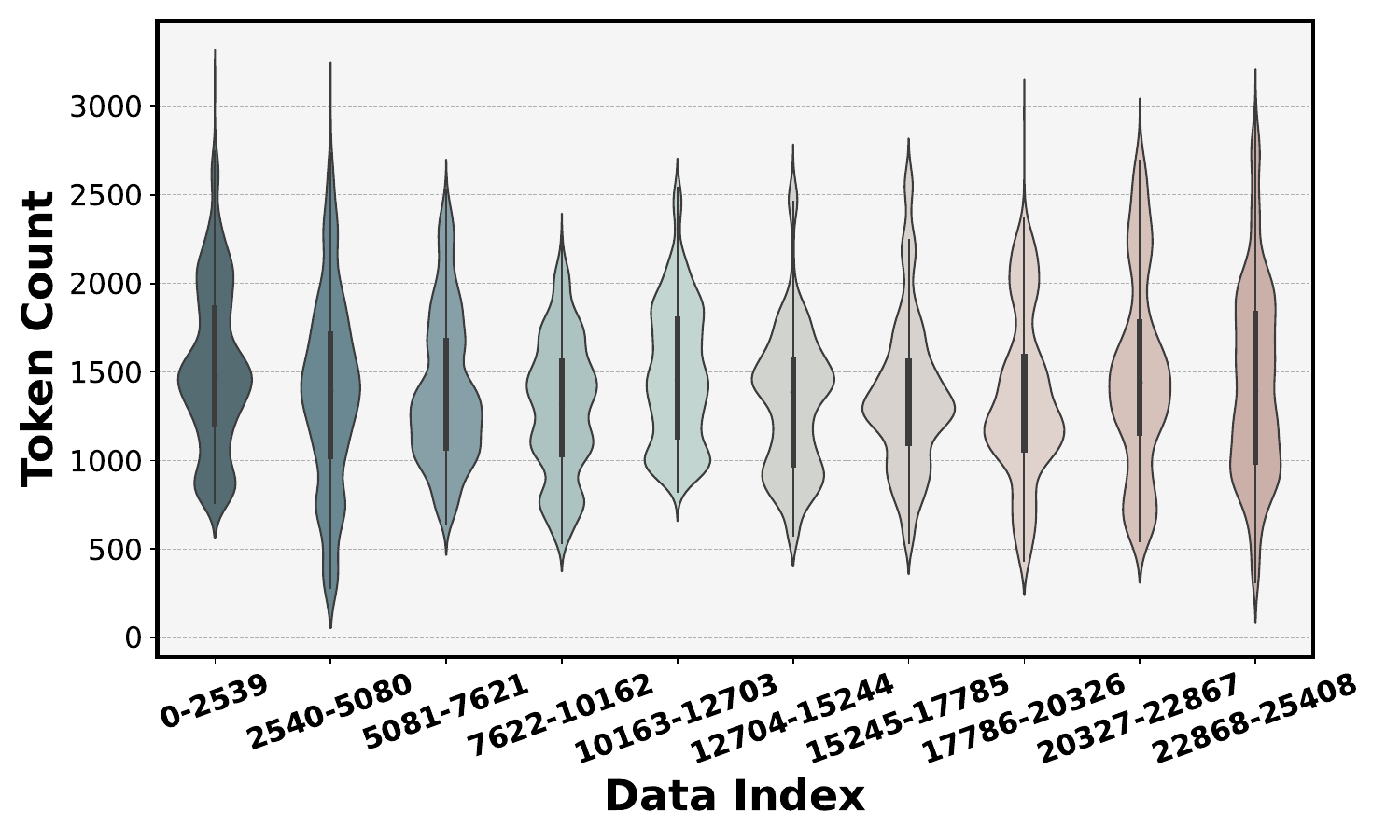}
        \caption{Premise token length distributions across different data index groups.}
        \label{fig:token_dist}
    \end{subfigure}
    \hfill
    \begin{subfigure}[b]{0.28\textwidth}
        \centering
        \includegraphics[width=\textwidth]{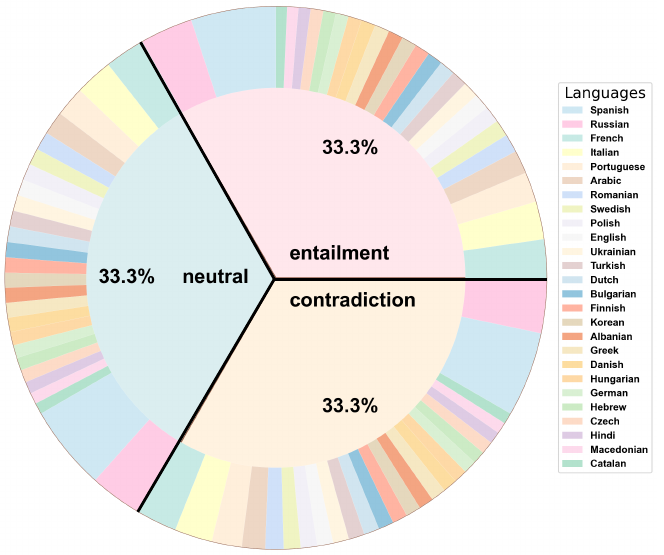}
        \caption{Label distribution with language composition.}
        \label{fig:label_dist}
    \end{subfigure}
    \caption{Statistic visualization of token length, EDU numbers, label distribution and language composition.}
    \label{fig:combined_analysis}
\end{figure*}
The hypothesis $\textit{H}$ is a sentence-level statement. The task requires reasoning over the combined information from $\textit{P}$ with $ H $ to determine their entailment relationship, involving both cross-document and cross-lingual premise integration.
\subsection{Premise Data Collection}
We collect our premises from GlobeSumm~\cite{ye2024globesumm}, a multi-document cross-lingual summarization dataset covering 370 topics across 26 languages. In GlobeSumm, documents for each topic span diverse media outlets, publication times, and languages, providing a rich foundation for cross-document and cross-lingual inference tasks. 
We curated CDCL-NLI dataset by stratified randomly selecting documents for each topic to form premise pairs. To enhance cross-lingual coverage, we strategically expanded our document collection through translation. **To address the cross-lingual aspect of the task, we used the DeepL API to translate the original English documents from GlobeSumm into 25 target languages. This translation process ensures consistent, high-quality multilingual premises.** After rigorous quality filtering, our final dataset consists of high-quality inference instances covering 26 languages. Detailed premise establishment criteria and quality filtering standards are provided in Appendix \ref{appendix:premise_criteria}.

\subsection{Hypothesis Generation and Label Specification}
For each pair, we generate hypotheses across three NLI categories. Initial hypotheses are generated by GPT-4o~\cite{openai2024gpt4} following specific guidelines \cite{wang-etal-2024-knowledgesg} to ensure balanced label distribution and sufficient reasoning depth. 
Entailment hypotheses require joint or consistent support from the premise documents.
Neutral hypotheses are plausible but neither supported nor contradicted.
Contradiction hypotheses explicitly conflict, focusing on cross-document inconsistencies. 
To reduce hallucination, GPT-4o first generates explanations before finalizing hypotheses.
Detailed prompts and protocols are included in Appendix \ref{app:nli_label_defination_hypo_gen}.

\subsection{Manual Annotation and Quality Control}
Our annotation involved two phases: hypothesis verification and EDU-based explanation (Figure~\ref{fig:data}). All human annotation was conducted on the original English versions of the premises and hypotheses. This design choice ensures that annotators did not require multilingual capabilities, and it minimizes the language gap during the critical verification process. To assess inter-annotator agreement, we randomly divided our training data into three equal parts. Each part was independently annotated by two of our three graduate students. This setup allowed us to calculate Cohen's $\kappa$ for each of the three annotator pairs, yielding an average $\kappa$ of 0.71 across these pairs, which indicates strong agreement. For explanations, annotators selected minimal EDU sets supporting their decisions, with high agreement (Jaccard: 0.91; span overlap: 0.94; conclusion: 1.00). All annotations were reconciled through discussions to ensure quality (see Appendix~\ref{app:annotation}). 
The final dataset contains multilingual premise-hypothesis pairs, NLI labels, and EDU node indices for explanation, with clear metadata indicating the source of each document.

\begin{table}[t]
\centering
\small
\begin{tabular}{lccccc}
\toprule
\textbf{Dataset} & \textbf{CD} & \textbf{CL} & \textbf{Interp.} & \textbf{Avg.Tks} & \textbf{Labels} \\
\midrule
MultiNLI     & \(\times\) & \(\times\) & \(\times\) & 33.7 & 3  \\
XNLI        & \(\times\) & \(\checkmark\) & \(\times\) & 50 & 3 \\
e-SNLI      & \(\times\) & \(\times\) & \(\checkmark\) & 45.1 & 3 \\
DocNLI      & \(\checkmark\) & \(\times\) & \(\times\)& 412 & 2 \\
\midrule
CDCL-NLI    & \(\checkmark\) & \(\checkmark\) & \(\checkmark\) & 1,456 & 3\\
\bottomrule
\end{tabular}
\caption{Characteristics of NLI datasets showing cross-document (CD), cross-lingual (CL), and interpretability (Interp.) capabilities, along with average tokens per instance (Avg.Tks) and number of label classes.}
\label{tab:dataset_stats}

\end{table}
    
% \begin{table}[h]
% \small
% \caption{Dataset statistics. All splits maintain balanced class distribution (33.3\% for each of Entailment, Neutral, and Contradiction).}
% \label{tab:dataset_stats}
% \begin{tabular}{l|c|c|ccc}
% \toprule
% \multirow{2}{*}{Split} & \multirow{2}{*}{Inst.} & \multirow{2}{*}{Topics} & \multicolumn{3}{c}{Avg} \\
% \cmidrule{4-6}
% & & &EDUs & Pre.Tks & Hyp.Tks \\
% \midrule
% Train & 22200 & 296 & 92 & 1,420 & 36 \\
% Dev & 1605 & 37 & 92 & 1,417 & 38 \\
% Test & 1605 & 37 & 93 & 1,429 & 38 \\
% \midrule
% Total & 25410 & 370 & --- & --- & --- \\
% \bottomrule
% \end{tabular}
% \footnotesize{
% Inst.: instances; Pre.Tks: premise tokens; Hyp.Tks: hypothesis tokens; EDUs: Elementary Discourse Units.
% }
% \end{table}

\subsection{Dataset Statistics}
We summarize the key characteristics of different NLI datasets in Table~\ref{tab:dataset_stats}, which shows substantial variations in their cross-document and cross-lingual capabilities.
Our CDCL-NLI dataset consists of 25,410 cross-document, cross-lingual NLI instances spanning 26 languages and 370 events. We partitioned the dataset by event topics, yielding 22,200/1,605/1,605 train/dev/test instances with mutually exclusive event distributions.
Figure~\ref{fig:data_split} shows similar data characteristics across training, validation, and test sets; Figure~\ref{fig:token_dist} depicts token count variations across consecutive segments; and Figure~\ref{fig:label_dist} illustrates balanced label distributions (33.3\% each) with roughly uniform language distribution within each label.
We provide more information about our dataset in Appendix \ref{app:data_info}.

\vspace{-2mm}

\section{Our Method: RST-enhanced Graph Fusion with EDU Level Interpretability}
Our approach offers a robust solution for cross-document and cross-lingual NLI by leveraging RST-enhanced graph fusion and explanation prediction. As illustrated in Figure~\ref{fig:model}, the framework comprises three main components: RST graph construction and fusion module, graph representation generation module and interpretability and classification module.

We employ DM-RST~\cite{liu2021dmrst} parser for discourse modeling as it offers an optimal balance between structural richness and computational feasibility. Compared to the locally-focused PDTB~\cite{prasad2008pdtb}, RST's hierarchical structure effectively captures document-level organization essential for cross-document reasoning. While SDRT~\cite{asher2003logics} is semantically richer, its $O(n^3)$ complexity is prohibitive for large-scale tasks. 
Our ablation study (Table \ref{tab:main_results}) empirically validates the effectiveness of our RST parser, showing that including the RST graph module can significantly improve performance despite potential parsing errors.

% We acknowledge that the performance of our DM-RST parser~\cite{liu2021dmrst} is lower than that of syntactic parsers, which reflects the inherent complexity of document-level analysis. However, our ablation studies (Table 3) demonstrate that removing the RST graph module causes a significant performance drop, confirming that the RST structure provides critical discourse information and a tangible gain despite potential parsing errors.

\subsection{RST Graph Construction and Fusion}

\begin{figure*}
  \centering
  \includegraphics[width=\linewidth]{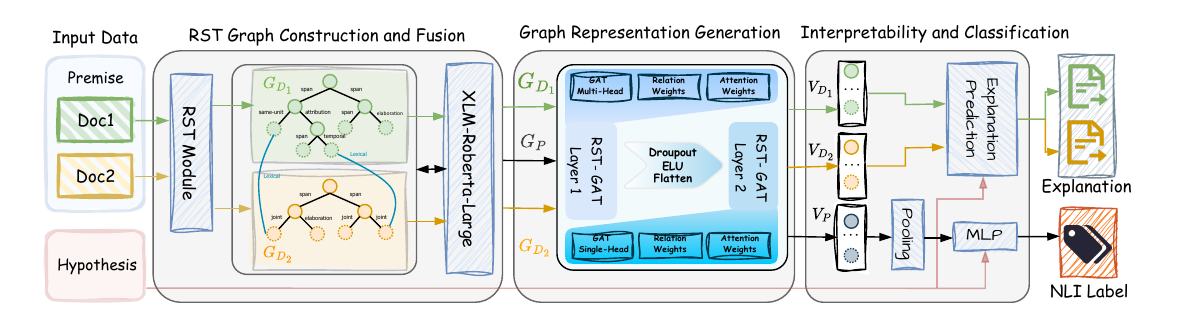}
  \vspace{-5mm}
    \caption{Our CDCL-NLI framework processes premise documents (${D_1,D_2}$) and a hypothesis through a multi-stage process: 1) \textbf{RST Graph Construction}, where an RST parser generates initial discourse structures (${G_{D_1}}$ and ${G_{D_2}}$) which are then fused into a single premise graph (${G_P}$) using semantic edges derived from XLM-RoBERTa embeddings; 2) \textbf{Graph Representation}, where the fused graph is processed by \textit{RST-GAT} layers; and 3) \textbf{Interpretability and Classification}, which extracts node-level explanations and uses the graph representations ($\bm{h}_{G_p}$) and hypothesis representation ($\bm{h}_{hypo}$) to predict the final NLI label.}
  \label{fig:model}

\end{figure*}
\begin{figure}[t]
    \centering
    \includegraphics[width=\linewidth]{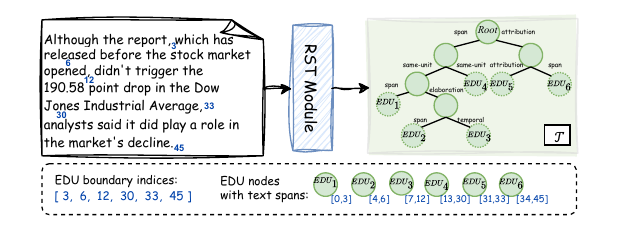}
    \vspace{-2mm}
    \caption{RST graph construction. The RST module first segments text into EDUs(EDU$_1$-EDU$_6$), with boundaries in blue, and then organizes an RST tree $\mathcal{T}$ showing discourse relations.}
    \label{fig:rst_example}

\end{figure}
\paragraph{RST Information Extraction.}
\label{subsubsec:RST Information Extraction}
% The RST module first segments the raw text into EDUs, with segmentation boundary indices shown in green. The segmentation yields corresponding EDU nodes (EDU\textsubscript{1} to EDU\textsubscript{6}) with their text spans. These EDUs are then organized into an RST tree structure that captures discourse relations.
We employ DM-RST~\cite{liu2021dmrst}, a top-down multilingual document-level rhetorical structure parsing framework, to extract RST information from the premise documents. As shown in Figure~\ref{fig:rst_example}, DM-RST generates two key features for document $D$: 1) EDU boundary indices and 2) RST tree parsing outputs. By processing these features, we get $ \textit{D} = \{EDU_1, EDU_2, ..., EDU_n\}$ and rhetorical structure tree $\mathcal{T}$. ${EDU_i}$ represents the $i\text{-th}$ EDU's textual content. $\mathcal{T}$ is formally defined as:
\vspace{-2mm}
\begin{displaymath}
\mathcal{T} = \left\{ 
\begin{aligned}
& (EDU_{[s \to t]}, EDU_{[t+1 \to u]}, r_{st}, r_{tu}) \mid \\
& s, t, u \in [1, n], \, s \leq t < u, \, r_{st},r_{tu} \in \mathcal{R} 
\end{aligned}
\right\},
\end{displaymath}
where $EDU_{[s \to t]}$ denotes an EDU group that forms either a leaf node (when $s=t$) or a branch node (when $s<t$), and $r_{st}$ represents the rhetorical relation. This tree structure captures both local EDU relationships and global discourse organization.

\paragraph{Embedding Model.}
\label{subsubsec:Embedding model}
To handle inconsistent cross-lingual encoding from premise documents in different languages, we use XLM-RoBERTa-Large~\cite{Conneau2020uclr} as the base encoder, which supports over 100 languages and excels at multilingual semantic representation. For each $EDU_i$ in the RST structure, its initial vector is $\bm{h}_{EDU_i} = \phi(EDU_i) \in \mathbb{R}^d$, where $\phi$ denotes XLM-RoBERTa-Large and $d=1024$. The hypothesis vector $\bm{h}_{\mathrm{hypo}}$ is computed similarly.
\paragraph{Single Graph Construction.}

Based on the RST tree $\mathcal{T}$, we construct graphs $G_{D_1}$ and $G_{D_2}$ for each document \(D_1\) and \(D_2\) respectively as shown in Figure~\ref{fig:model}. For graph $G({V}, {E}, {R})$, we define:

\begin{itemize}[nosep, left=0em]
  \item \textbf{Node Set} $\mathcal{V} = \{ v_i \mid \text{EDU}_{[s \to t]} \in \mathcal{T} \}$, where each $v_i$ has features: $\text{Text}_{v_i}$, $\phi_{v_i}$, and $\text{Type}_{v_i}$ (e.g., nucleus or satellite).
  \item \textbf{Edge Set} $\mathcal{E} = \{(v_i,v_j) \mid v_i \neq v_j, (v_i,v_j,r) \in \mathcal{T}\}$, representing typed, bidirectional edges with rhetorical relations.
  \item \textbf{Relation Set} $\mathcal{R}$ is from rhetorical relations in $\mathcal{T}$.
\end{itemize}
For detailed relations and definitions of node features, please refer to the Appendix \ref{app:graph_relation_types}, \ref{app:graph node feature}.

\paragraph{Graph Fusion.}
After obtaining heterogeneous graphs $G_{D_1}(V_{D_1}, E_{D_1}, R)$ and $G_{D_2}(V_{D_2}, E_{D_2}, R)$ for the premise, we then merge them via lexical chains to enhance cross-document reasoning by:

\begin{itemize}[nosep, left=0em]
    \item \textbf{Node Feature Fusion}: $V_P = V_{D_1} \cup V_{D_2}$, retaining all nodes and features.
    \item \textbf{Cross-document Edge}: Add bidirectional lexical edges between $v_i \in V_{D_1}$ and $v_j \in V_{D_2}$ if $\mathrm{CosineSim}(v_i, v_j) > \delta$, and obtain \( E_P\).\footnote{Threshold \(\delta\) is chosen empirically; see Appendix~\ref{app:threshold} for detailed justification.}
    \item \textbf{Adding Edge Types}: Extend $R$ with a new "Lexical" relation $R'$ to support lexical alignment.
\end{itemize}
The merged graph $G_P(V_P, E_P, R')$ preserves individual features while aligning semantics across documents, effectively supporting CDCL-NLI.

% \subsection{Graph Processing}
\subsection{Graph Representation Generation}
\label{subsec:Graph Processing}
\paragraph{Node-level Representation.}
As shown in Figure~\ref{fig:model}, there are two layers of \textit{RST-GAT} to process nodes' features. 
\textit{RST-GAT} builds upon the Relation-aware Graph Attention Network (RGAT)~\cite{busbridge2019rgat}, which extends Graph Attention Network (GAT)~\cite{velickovic2018gat} to handle relation-specific edge types in graphs.

Taking a graph $G(V, E, R)$ as an example, the initial node embeddings $\mathbf{h}_V^0$
% $\mathbf{h}_V^0 = \{\mathbf{h}_{v_i}^0 = \phi(v_i)\}$ 
are obtained as described in Section~\ref{subsubsec:Embedding model}. Node representations are then updated through two layers of relation-aware multi-head attention as follows:
\vspace{-2mm}
\begin{equation}
\label{eq:graph_attention}
\scalebox{0.85}{$
\mathbf{h}_{v_i}^{(l)} = \frac{1}{|R|} \sum\limits_{r \in R} \alpha_r \cdot \frac{1}{K} \sum\limits_{k=1}^K \sum\limits_{v_j \in \mathcal{N}_r(v_i)} \beta_{ij,k}^{r,(l)} \mathbf{W}_{r,k} \mathbf{h}_{v_j}^{(l-1)}
$}
\end{equation}
where $l = 1, 2$. Here, $\alpha_r$ denotes the softmax-normalized weight of relation $r$, capturing the relative importance among relations, while $\beta_{ij,k}^{r,(l)}$ represents the attention coefficient over neighboring nodes, indexed by node pairs $(v_i, v_j)$, attention head $k$, relation $r$, and layer $l$. 
After two layers of message passing, the resulting node embeddings are denoted as $\mathbf{h}_V = \{\mathbf{h}_{v_i}^{(2)}\}$. 
The same update procedure is applied independently to $G_{D_1}$, $G_{D_2}$, and $G_P$, producing embeddings $\mathbf{h}_{V_{D_1}}, \mathbf{h}_{V_{D_2}},$ and $\mathbf{h}_{V_P}$, respectively. 
Detailed formulations of the attention weights and parameter configurations are provided in Appendix~\ref{app:attention}.

\paragraph{Graph-level Representation.}

The global representation($\bm{h}_{G_P}$) of the merged graph $G_P$ is obtained by averaging node features after two \textit{RST-GAT} layers. 
% Specifically, we compute $\bm{h}_{G_P} = \frac{1}{N} \sum_{i=1}^N \bm{h}_{v_i}$, where $\bm{h}{v_i} \in \bm{h}_{V_P}$, $i$ means the $i\text{-th}$ node in $G_P$, and $N$ is the number of nodes in $G_P$. 
This pooling captures discourse-level semantics while preserving local rhetorical relations, enabling effective classification.

\paragraph{Classification Loss.}
Given the concatenated graph representation $\bm{h}_{G_p}$ and hypothesis features $\bm{h}_{hypo}$, the classification loss is computed using the standard cross-entropy (CE) formulation:
\begin{equation}
\scalebox{0.82}{$
\mathcal{L}_{\mathrm{cls}} = \mathrm{CE}(\mathbf{y}, \mathrm{Softmax}(\mathrm{MLP}(\bm{h}_{G_p} \oplus \bm{h}_{hypo})) \in \mathbb{R}^3),
$}
\end{equation}
where $\mathbf{y}$ denotes the ground-truth label and $\mathbf{p}$ denotes the predicted probability distribution.

\paragraph{Enhanced Triplet Loss.}
Triplet loss~\cite{weinberger2006distance, schroff2015facenet} is a metric learning method that encourages the anchor-positive distance to be smaller than the anchor-negative distance. Leveraging the structure of our CDCL-NLI dataset, where each premise aligns with three hypotheses (entailment, neutral, contradiction), we propose a neutral-constrained triplet loss:
\begin{equation}
\scalebox{0.9}{
$\begin{aligned}
\mathcal{L}_{\text{triplet}} = \max(0, d(a, p) - d(a, n) + \sigma)
\\
+ \max(0, d(a, \text{neu}) - d(a, n) + \theta),
\end{aligned}$
}
\end{equation}
where $d(x, y)$ is the Euclidean distance, and $a, p, \text{neu}, n$ denote the premise paired with entailment, neutral, and contradiction hypotheses, respectively. Margins $\sigma$ and $\theta$ enforce the semantic order: entailment $<$ neutral $<$ contradiction.
% Leveraging our dataset structure with entailment, neutral, and contradiction hypotheses, we define:
% \begin{equation}
% \scalebox{0.9}{
% $\begin{aligned}
% \mathcal{L}_{\mathrm{triplet}} = {} & \max(0, d(a, p) - d(a, n) + \sigma) \\
% & + \max(0, d(a, u) - d(a, n) + \theta)
% \end{aligned}$
% }
% \end{equation}
% where $d(\cdot,\cdot)$ is Euclidean distance, $a$, $p$, $\mathrm{neu}$, $n$ represent premise and corresponding hypotheses, and $\theta$, $\sigma$ are margins.
\subsection{EDU-level Explanation Prediction}

For interpretability, we propose an attention-based method to extract explanation nodes.
\paragraph{Node Importance.}  
Using multi-head attention weights from the first \textit{RST-GAT} layer, the importance score \(I_i\) of node \(v_i\) in \(G_{D_1}, G_{D_2}\) is
\vspace{-1mm}
\begin{equation}
\scalebox{0.85}{$
I_i = \frac{1}{K} \sum_{k=1}^K \sum_{r \in R} \sum_{v_j \in \mathcal{N}_r^{\mathrm{in}}(v_i)} \beta_{ji,k}^{r,(1)}.
$}
\end{equation}
Let \(\bm{H} = [\bm{h}_{v_0}; \ldots; \bm{h}_{v_n}] \) be node features and \(\bm{I} = [I_0, \ldots, I_n]^\top \) importance scores. Weighted features are \(\bm{H}' = \bm{I} \odot \bm{H}\), where \(\odot\) denotes element-wise product with broadcasting.

\paragraph{Hypothesis-aware Interaction.}  
Given hypothesis embedding \(\bm{h}_{hypo} \in \mathbb{R}^{d^{\mathrm{out}}}\), attention over weighted features \(\bm{H}' \in \mathbb{R}^{n \times d^{\mathrm{out}}}\) produces interaction features:

\vspace{-2mm}
\begin{equation}
\scalebox{0.85}{$
\bm{O} = \mathrm{Attention}\left(\frac{\bm{h}_{hypo} \bm{H}'^\top}{\sqrt{d^{\mathrm{out}}}}\right) \bm{H}'.
$}
\end{equation}

\paragraph{Feature Fusion and Classification.}
The model is optimized by Binary Cross-Entropy (BCE) loss:
\vspace{-2mm}
\begin{equation}
\scalebox{0.83}{$\displaystyle
\mathcal{L}_{\mathrm{exp}} = \frac{1}{N} \sum_{i=1}^N \mathrm{BCE}\bigl(y_i, \text{Sigmoid}(\mathrm{MLP}([\bm{h}'_i \oplus \bm{o}_i]))\bigr)
$}
\end{equation}
where $y_i \in \{0,1\}$ is ground truth label of node $i$, $\bm{h}'_i$ and $\bm{o}_i$ are the weighted and interaction features for node $i$ respectively. 
% Nodes with $\hat{y}_i \geq 0.5$ are selected to form the explanation set $\mathcal{E}$ containing the corresponding text segments..

The total loss combines all components:
\begin{equation}
\scalebox{0.9}{$
\mathcal{L}_{\mathrm{total}} = \gamma \mathcal{L}_{\mathrm{exp}} + \lambda (\mathcal{L}_{\mathrm{cls}} + \mathcal{L}_{\mathrm{triplet})},
$}
\end{equation}
where $\gamma$ and $\lambda$ are balancing hyperparameters set as 0.2 and 0.8 respectively through grid search on the validation set.
\section{Experiments}

\subsection{Experiment Settings}

\begin{table*}[t]
\centering
\small
\setlength{\tabcolsep}{7pt}
\begin{tabular}{c|c|ccc|ccc|c}
\toprule
\multirow{2}{*}{\textbf{Model Type}} & \multirow{2}{*}{\textbf{Model}} & \multicolumn{3}{c|}{\textbf{TestSet1:Cross-Lingual}} & \multicolumn{3}{c|}{\textbf{TestSet2:English}} & \multirow{2}{*}{\textbf{Trained}}  \\
% \cmidrule(r){3-5} \cmidrule(l){6-8}
~ & ~ & Precision & Recall & F1 Macro & Precision & Recall & F1 Macro &  \\
\midrule
\multirow{3}{*}{\makecell{Conventional\\Model}} & Hypothesis-only & 35.78 & 36.02 & 35.84 & 35.89 & 35.97 & 35.91 & \checkmark \\
~ & DocNLI & 64.75 & 64.30 & 64.46 & 69.29 & 68.39 & 68.70 & \checkmark \\
~ & R2F & 65.04 & 65.42 & 65.42 & 67.18 & 68.47 & 67.13 & \checkmark \\
\midrule
\multirow{3}{*}{\makecell{Large\\Language\\Model}} & Llama-3-8B  & 45.94 & 52.62 & 48.07 & 51.69 & 57.98 & 53.03 & \checkmark \\
~ & GPT-4o & 52.50 & 56.30 & 54.00 & 62.50 & 65.00 & 64.50 & $\times$ \\
~ & Qwen3-8B & 60.34 & 56.29 & 59.86 & 71.71 & 67.62 & 67.34 & \checkmark  \\
\midrule
\multirow{4}{*}{\makecell{CDCL-NLI\\Model}} & Ours & \textbf{71.09} & \textbf{70.84} & \textbf{68.95} & \textbf{72.65} & \textbf{72.46} & \textbf{70.68}  & \checkmark \\
~ & - Exp & 65.99 & 67.29 & 65.86 & 69.01 & 69.97 & 68.79 & \checkmark \\
~ & - Graph & 53.07 & 57.38 & 51.37 & 68.64 & 64.55 & 61.71 & \checkmark \\
~ & - Exp \& Graph & 49.15 & 52.71 & 48.70 & 49.15 & 52.71 & 50.67 & \checkmark \\
\bottomrule
% \multicolumn{8}{l}{\textsuperscript{†}Few-shot setting} \\
\end{tabular}

\caption{NLI model performance on cross-lingual (TestSet1) and English (TestSet2) sets. Our full model achieves the highest F1 scores, showing clear gains from explanation and graph components. Large language models perform well but are generally outperformed. \checkmark~indicates training on target data; $\times$ means no training. Explanation - Exp.}
\label{tab:main_results}

\end{table*}
% \begin{table*}[htbp]
% \centering
% \small
% \caption{Performance comparison on CDCL-NLI dataset. Best results are in \textbf{bold}.}
% \label{tab:main_results}
% \begin{tabular}{l|c|ccc|ccc|c}
% \toprule
% \multirow{2}{*}{\textbf{Model Type}} & \multirow{2}{*}{\textbf{Model}} & \multicolumn{3}{c|}{\textbf{Dev}} & \multicolumn{3}{c|}{\textbf{Test}} & \multirow{2}{*}{\textbf{Trained}}  \\
% % \cmidrule(r){3-5} \cmidrule(l){6-8}
%  &  & Precision & Recall & F1 Macro & Precision & Recall & F1 Macro &  \\
% \midrule
% & Hypothesis-only & 36.12 & 35.89 & 35.97 & 35.78 & 36.02 & 35.84 & \checkmark \\
% Conventional& DocNLI & 68.67 & 67.54 & 67.86 & 64.75 & 64.30 & 64.46 & \checkmark \\
% Model& R2F & 65.23 & 65.67 & 65.28 & 65.04 & 65.42 & 65.42 & \checkmark \\
% \midrule
% Large & Llama-3-8B\textsuperscript{†} & 65.42 & 64.86 & 64.21 & 64.38 & 63.77 & 63.20 & \checkmark \\
% Language & GPT-4o\textsuperscript{†} & 67.13 & 66.50 & 66.00 & 65.50 & 65.00 & 64.50 & $\times$ \\
% Model & Qwen3-8B\textsuperscript{†} & 69.45 & 68.92 & 67.85 & 68.24 & 67.10 & 66.78 & \checkmark \\
% \midrule
% & Ours & \textbf{72.65} & \textbf{72.46} & \textbf{70.68} & \textbf{71.09} & \textbf{70.84} & \textbf{68.95} & \checkmark \\
% Ours & - Exp & 69.01 & 69.97 & 68.79 & 65.99 & 67.29 & 65.86 & \checkmark \\
% Model & - Graph & 68.64 & 64.55 & 61.71 & 53.07 & 57.38 & 51.37 & \checkmark \\
% & - Exp \& Graph & 49.15 & 52.71 & 53.29 & 49.15 & 52.71 & 48.70 & \checkmark \\
% \bottomrule
% \multicolumn{8}{l}{\textsuperscript{†}Few-shot setting} \\
% \end{tabular}
% \end{table*}

\paragraph{Metrics.}
Model evaluation considers classification and explanation quality. For classification on DocNLI (imbalanced), we report \textbf{Micro F1} and \textbf{Weighted F1}. On CDCL-NLI dataset, we use \textbf{Macro Precision}, \textbf{Macro Recall}, and \textbf{Macro F1} for balanced class performance. Explanation quality is assessed using \textbf{BLEU} (1-4), \textbf{ROUGE-1/2/L}, and \textbf{METEOR}.

% We evaluate the model from two perspectives: classification performance and explanation quality. For classification, we report \textbf{Micro F1} and \textbf{Weighted F1} on the DocNLI dataset to account for its highly imbalanced label distribution. On the CDCL-NLI dataset, we focus on \textbf{Macro Precision}, \textbf{Macro Recall}, and \textbf{Macro F1} to reflect the unweighted average performance across classes. For explanation quality, we use \textbf{BLEU} (1- to 4-gram overlap) alongside \textbf{ROUGE-1}, \textbf{ROUGE-2}, \textbf{ROUGE-L} and METERO metrics to comprehensively assess the similarity between generated and reference explanations.

% \paragraph{Implementation Details.}
% Our model is implemented in PyTorch and trained on an NVIDIA A100 GPU. Key hyperparameters are as follows: the learning rate is set to $1 \times 10^{-5}$ with a cosine warmup schedule, and the batch size is 20. We use the AdamW optimizer with a weight decay of $1 \times 10^{-5}$, and the model is trained for 20 epochs. 

\paragraph{Baselines.}
\begin{itemize}[nosep, left=0.5em]
    \item \textbf{Conventional NLI Models}: We compare two well-established models, both trained on our dataset: \textbf{DocNLI}~\cite{yin2021docnli}, a document-level NLI model tailored for long texts, and \textbf{R2F}~\cite{wang2022r2f}, a retrieval-based framework for document-level NLI. All conventional baselines and our proposed method are built upon the same underlying pretrained language model to ensure fair comparison. Training details are provided in Appendix~\ref{app:conventional hyperparameters}.
    \item \textbf{Large Language Models}: We evaluate three LLMs: \textbf{Llama3-8B-Instruct}~\cite{meta2024llama3}, \textbf{Qwen-3-8B}~\cite{qwen3} and \textbf{GPT-4o}~\cite{openai2024gpt4}, where the LLaMA and Qwen model is further fine-tuned with LoRA adapters. All models are tested in a few-shot setting, with fine-tuning configurations in Appendix~\ref{app:LLM Details}.
\end{itemize}

\subsection{Experiment Results and Analysis}

\paragraph{Main Results and Ablation Study.}
Table~\ref{tab:main_results} presents a performance comparison of our proposed method against several competitive baselines on two test sets. TestSet1 is a cross-lingual test set (the original test set of the CDCL-NLI dataset). TestSet2 is an English-translated version of TestSet1, designed to evaluate model robustness in a cross-document scenario without language barriers, and to quantify the performance degradation caused by cross-lingual factors. This dual evaluation framework enables a clearer analysis of the impact of language variation on NLI performance.\footnote{Unless noted, all reported test results refer to TestSet1.}

Our model consistently achieves the best results on both test sets, with macro F1 scores of \textbf{68.95\%} on the cross-lingual set and \textbf{70.68\%} on the English-translated set, surpassing strong baselines such as DocNLI and R2F by notable margins. The generally higher scores on the English test set highlight the relative ease of reasoning within a single, well-resourced language, in contrast to the added challenges of cross-lingual understanding, which requires effective language transfer and alignment. The hypothesis-only baseline, which trains solely on the hypothesis, attains near-random performance (~36\% F1), indicating minimal dataset artifacts in the hypothesis statements.

Among the large language models evaluated in the few-shot setting, Qwen3-8B achieves the best performance, with F1 scores of 59.86\% on the cross-lingual set and 67.34\% on the English set, outperforming both GPT-4o and Llama3-8B. Nevertheless, our approach surpasses Qwen3-8B by 9.09\% on the cross-lingual set and 3.34\% on the English set, highlighting the effectiveness of our method. Detailed prompts and zero-shot results and reported in Appendix~\ref{app:evaluation_prompt},  Appendix~\ref{app: LLM_zero}.

The ablation study highlights the importance of each component: removing the explanation module (- Exp) results in a moderate performance drop of 1.89\% on both cross-lingual and English test sets; removing the graph module (- Graph) causes a more pronounced decline of 17.58\% and 8.97\%, respectively. When both components are removed (- Exp \& Graph), performance sharply decreases on both test sets, demonstrating that these modules jointly contribute to the model’s robustness under different language conditions.

\paragraph{Single-Document vs Cross-Document.}
\begin{figure}[t]
\centering
\includegraphics[width=0.8\linewidth]{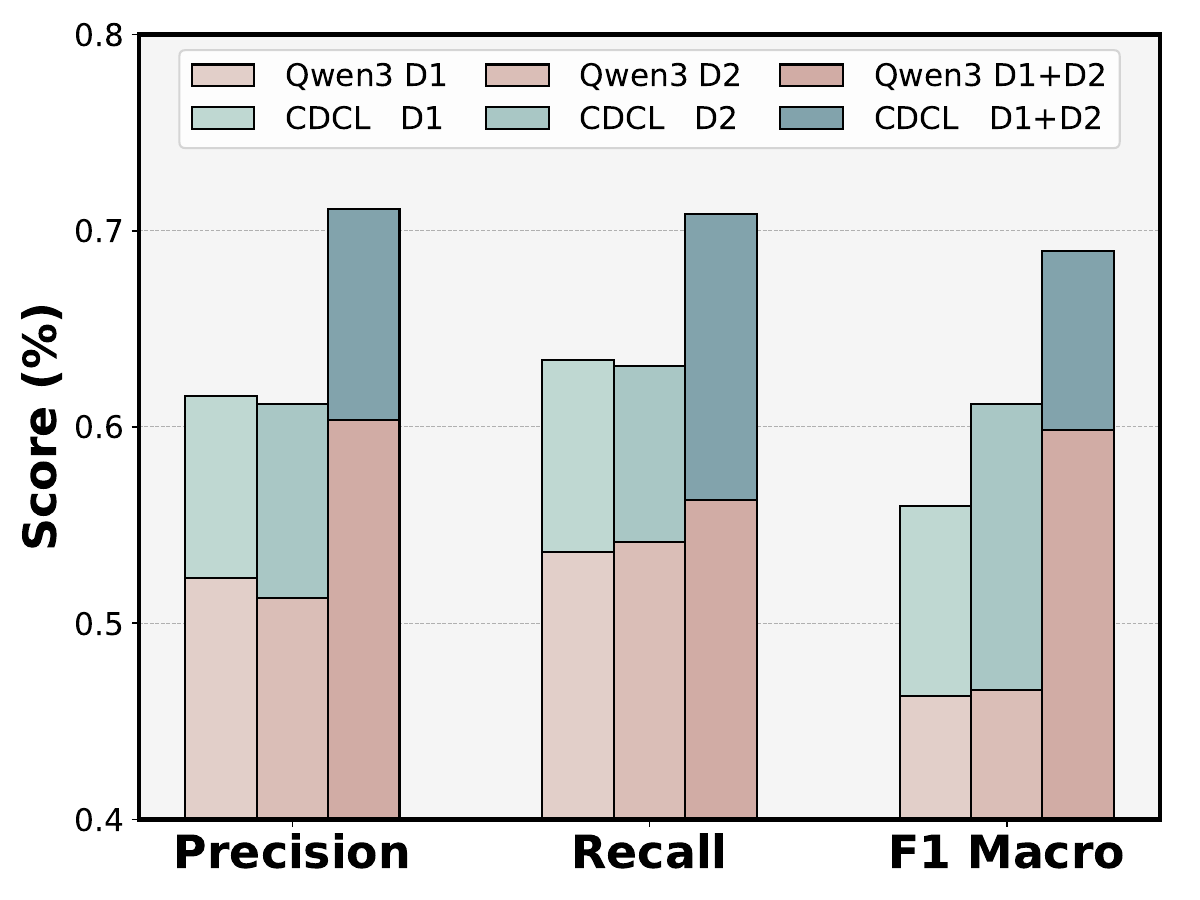}
\caption{NLI performance using single documents ($D1$, $D2$) versus combined ($D1+D2$). The F1 gain confirms the need for cross-document reasoning, with both documents contributing similarly.}
% \caption{Comparison of NLI performance using single-document premises ($D1$ or $D2$) versus combined documents ($D1+D2$). The notable $\sim$5\% F1 score gain when using both documents confirms the necessity of cross-document reasoning. Similar scores for $D1$ and $D2$ indicate both contribute equally to inference.}
\label{fig:single_doc}

\end{figure}
To validate the cross-document nature of our dataset, we compare the performance of models using only a single document ($D1$ or $D2$) against those using the $D1+D2$, as illustrated in Figure \ref{fig:single_doc}. The substantial performance gap—at least a 7\% F1 improvement—demonstrates that effective inference requires integrating information from both documents. Additionally, the similar F1 scores for $Document_1$ (63.2\%) and $Document_2$ (62.8\%) indicate that both documents provide equally important information, underscoring the necessity of synthesizing evidence from both sources rather than relying on either alone. Additional results are presented in Appendix \ref{app: baseline_singledoc}.
% The consistent pattern across both development and test sets, where full-premise models outperform single-document baselines by a similar margin, demonstrates the robustness of this cross-document dependency. 

\paragraph{Cross-Lingual Generalization.}
\begin{table}
\small
\begin{tabular}{@{}cccc@{}}
\toprule
\multicolumn{4}{c}{\textbf{F1 Scores on Target Language (Ours vs. R2F)}} \\
\midrule
\textcolor{lightblue}{ES}$\rightarrow$RU & \textcolor{lightblue}{ES}$\rightarrow$FR & \textcolor{lightblue}{ES}$\rightarrow$IT & \textcolor{lightblue}{ES}$\rightarrow$EN \\
\textbf{55.53}/25.03 & \textbf{58.28}/27.31 & \textbf{54.68}/29.31 & \textbf{57.94}/34.21 \\
\midrule
\textcolor{lavender}{RU}$\rightarrow$ES & \textcolor{lavender}{RU}$\rightarrow$FR & \textcolor{lavender}{RU}$\rightarrow$IT & \textcolor{lavender}{RU}$\rightarrow$EN\\
\textbf{52.83}/46.26 & \textbf{46.67}/35.50 & \textbf{50.89}/39.77 & \textbf{49.67}/47.78\\
\midrule
\textcolor{lightgreen}{FR}$\rightarrow$ES & \textcolor{lightgreen}{FR}$\rightarrow$RU & \textcolor{lightgreen}{FR}$\rightarrow$IT & \textcolor{lightgreen}{FR}$\rightarrow$EN \\
\textbf{50.31}/43.25 & \textbf{56.6}/22.24  & \textbf{58.65}/39.32 & \textbf{49.67}/47.22 \\
\midrule
\textcolor{lightyellow}{IT}$\rightarrow$ES & \textcolor{lightyellow}{IT}$\rightarrow$RU & \textcolor{lightyellow}{IT}$\rightarrow$FR & \textcolor{lightyellow}{IT}$\rightarrow$EN \\
\textbf{53.72}/36.01 & \textbf{57.19}/36.21 & \textbf{53.17}/37.22 & \textbf{56.67}/47.21 \\
\midrule
\textcolor{lightorange}{EN}$\rightarrow$ES & \textcolor{lightorange}{EN}$\rightarrow$RU & \textcolor{lightorange}{EN}$\rightarrow$FR & \textcolor{lightorange}{EN}$\rightarrow$IT \\
\textbf{60.31}/49.94 & \textbf{51.27}/32.46 & \textbf{60.28}/30.80 & \textbf{55.11}/38.33 \\
\bottomrule
\end{tabular}

\caption{Cross-lingual performances (macro F1 scores) of our method and R2F. Source languages are colored.
\textcolor{lightblue}{Spanish} (\textcolor{lightblue}{ES}),  \textcolor{lavender}{Russian} (\textcolor{lavender}{RU}), \textcolor{lightgreen}{French} (\textcolor{lightgreen}{FR}), \textcolor{lightyellow}{Italian} (\textcolor{lightyellow}{IT}) and \textcolor{lightorange}{English} (\textcolor{lightorange}{EN}).
Our method demonstrates superior generalization across languages compared to baselines.
}
\label{tab:cross_lang}

\end{table}

To further assess the robustness and generalization of our approach, we conduct cross-lingual transfer experiments in a challenging scenario where the training and testing languages are distinct. Specifically, we select five typologically and geographically diverse languages—Spanish, Russian, French, Italian, and English—to ensure comprehensive coverage and to reflect real-world multilingual settings. For each source language, we translate the data into all target languages, resulting in 20 transfer directions. Models are trained on one language and evaluated on a different target language, with no overlap between training and test languages. As shown in Table~\ref{tab:cross_lang}, our method consistently outperforms the R2F baseline across most transfer directions, often by substantial margins. R2F is chosen as it improves upon DocNLI for cross-document reasoning. These results demonstrate the effectiveness of our approach in synthesizing information from cross-lingual document pairs and its strong transferability to diverse language pairs, validating the design of our experimental setup and the broad applicability of our method in multilingual cross-document NLI tasks.

\paragraph{Interpretability Study.}

\begin{figure}[t]
\centering
\includegraphics[width=0.6\linewidth]{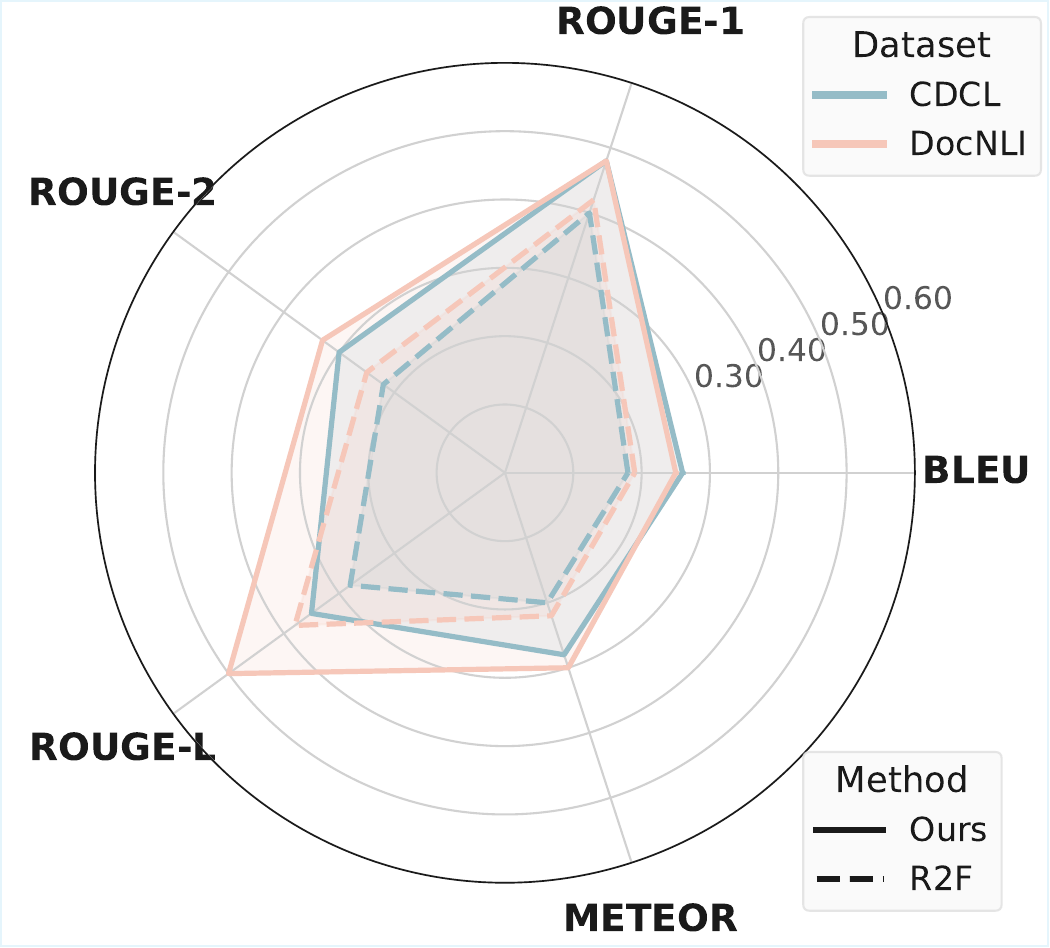}
\caption{Explainability comparison between our method and R2F on CDCL-NLI and DocNLI datasets using BLEU, ROUGE (1/2/L), and METEOR metrics. Our method consistently outperforms R2F across all metrics and datasets.}
\label{fig:interpretability}

\end{figure}

To evaluate our method's effectiveness, we compared it against the R2F baseline using five standard metrics (ROUGE-1/2/L, BLEU, METEOR) on both CDCL and DocNLI datasets. As shown in Figure~\ref{fig:interpretability}, our method (solid line) consistently outperforms r2f (dashed line) across all metrics on both datasets. The improvements are particularly pronounced in ROUGE-L, where our method achieves 0.34 versus 0.30 on CDCL-NLI and 0.50 versus 0.37 on DocNLI, demonstrating enhanced capability in preserving structural coherence. It is worth noting that the interpretability data for DocNLI was provided by R2F.

% Together, these results highlight our model’s comprehensive capability to generate text that is lexically precise, structurally coherent, and semantically meaningful across diverse datasets and evaluation dimensions.

\begin{table}
% \normalsize
\small
\setlength{\tabcolsep}{9pt}
\begin{tabular}{l|cc|cc}
\toprule
\multirow{2}{*}{\textbf{Method}} & \multicolumn{2}{c|}{\textbf{Dev}} & \multicolumn{2}{c}{\textbf{Test}} \\
~ & W. F1 & Mi. F1 & W. F1 & Mi. F1 \\
\midrule
DocNLI & 88.05 & 86.25* & 87.09 & 85.06* \\
R2F & 90.18* & \textbf{89.15} & 89.16* & 87.86 \\
Ours & \textbf{91.58} & 88.61 & \textbf{90.30} & \textbf{88.47} \\
\bottomrule
\end{tabular}

\caption{Performance comparison on the document-level DocNLI. Results marked with * are from our reproduction. Weighted F1 -W. F1, Micro F1 - Mi. F1}
\label{tab:generalization}

\end{table}

\paragraph{Comparison on DocNLI Dataset.}
We evaluate the generalization of our method on the DocNLI dataset using weighted and micro F1 metrics. As shown in Table~\ref{tab:generalization}, our approach achieves state-of-the-art weighted F1, outperforming both the DocNLI baseline and R2F, but slightly underperforms R2F on micro F1. This is mainly due to class imbalance between training and evaluation sets, and R2F’s advantage on the simpler reasoning tasks common in DocNLI, while our method is optimized for more complex reasoning. These results suggest that balanced sampling or improved adaptability could further boost performance.

%To evaluate the generalization capability of our approach, we conduct experiments on the DocNLI dataset using both weighted F1 and micro F1 metrics to address the significant class imbalance. As shown in Table~\ref{tab:generalization}, our method achieves state-of-the-art performance in weighted F1 on both development (91.58\%) and test sets (90.30\%), outperforming the DocNLI baseline and R2F*. However, on the development set's micro F1, our approach (88.61\%) performs slightly below R2F (89.15\%).

%This performance pattern can be attributed to two factors: First, the class distribution disparity between training (49.5\% vs. 50.5\%) and evaluation sets (12\% vs. 88\%) impacts our method's effectiveness. Second, while R2F excels at simpler reasoning tasks that dominate DocNLI, our method is optimized for complex reasoning scenarios. These findings suggest potential improvements through balanced sampling strategies or enhanced model adaptability.

\section{Conclusion}
This work systematically investigates CDCL-NLI, addressing key challenges in cross-document reasoning and multilingual understanding. 
We introduce a novel CDCL-NLI dataset spanning 26 languages and comprising 25,410 meticulously annotated instances.
And we propose an RST-enhanced graph fusion mechanism with explanation prediction.
Through extensive experiments and analyses, we demonstrate that our method effectively captures both structural and semantic information across documents and languages. Specifically, the RST-enhanced graph fusion mechanism and explanation prediction component not only improve model interpretability but also enhance performance, as validated by our ablation studies.
% Our empirical findings provide several key insights: 1) integrating rhetorical structure significantly improves the graph model’s ability to capture document-level discourse information; 2) reasoning cross-document is necessary and our method has strong cross-lingual reasoning capability; and 3) the EDU-level attribution method has a beneficial effect on classification and could generate explanations aligned with human reasoning.

\section*{Acknowledgments}
This work is supported by the National Key Research and Development Program of China (No. 2022YFB3103602), the National Natural Science Foundation of China (No. 62176187, No. 62202210). This work is also supported by the open project of Sichuan Provincial Key Laboratory of Philosophy and Social Science for Language Intelligence in Special Education (No. YYZN-2023-1).

\section*{Limitations}
% Despite the promising results, several limitations warrant attention. 
Our current framework is constrained to reasoning between pairs of documents, while real-world scenarios often involve multiple documents across diverse topics. 
% 改成在futurework 里用多文档
% Second, the current graph fusion mechanism could benefit from incorporating more sophisticated cross-document relationship modeling. 不提模型
% , such as CST-based approaches. 
% Third, the reliance on supervised explanation generation incurs substantial annotation costs, suggesting the need for efficient unsupervised alternatives. 
This limitation points to valuable directions for future research in multi-document multi-lingual inference.
\section*{Ethics Statement}

All data in our proposed dataset are collected from publicly available sources with respect for privacy and copyright. We have removed any personally identifiable information during preprocessing. The dataset is intended for research purposes only, and we advise users to be aware of potential biases present in the original data.

\appendix

\bibliography{custom}

\section{Dataset Details}
\label{appendix:dataset_construction}

\subsection{Premise Establishment Criteria}
\label{appendix:premise_criteria}

To ensure the quality and reliability of our CDCL-NLI dataset, we establish the following criteria for premise selection:

\begin{itemize}[nosep, left=0.5em]
    \item \textbf{Content Parallelism}: The document pairs must discuss the same topic while being naturally written in their respective languages, rather than being translations of each other. This ensures authentic cross-lingual reasoning scenarios.
    
    \item \textbf{Information Complementarity}: While maintaining topic consistency, documents in different languages should present complementary perspectives or details, enabling meaningful cross-document inference tasks.
    
    \item \textbf{Language Distribution}: Premise document pairs are randomly sampled from different languages to reflect real-world cross-lingual scenarios. Each pair must consist of documents in two distinct languages, ensuring the dataset captures authentic cross-lingual reasoning challenges.
\end{itemize}

These criteria ensure that our dataset captures genuine cross-lingual reasoning challenges while maintaining natural language expression across different languages.
\subsection{CDCL-NLI Label Definitions and Hypothesis Generation}
\label{app:nli_label_defination_hypo_gen}
\paragraph{Label Definitions.}
We define three inference labels for CDCL-NLI, considering various evidence distribution scenarios across documents:

\begin{itemize}[nosep, left=0em]
    \item \textbf{Entailment}: The hypothesis is supported when either:
    \begin{itemize}[nosep, left=0.5em]
        \item Evidence from both documents jointly supports the hypothesis through cross-document reasoning, or
        \item One document provides sufficient supporting evidence while the other document contains no contradicting information
    \end{itemize}
    In both cases, the conclusion must be logically derivable without requiring external knowledge.
    
    \item \textbf{Contradiction}: The hypothesis is contradicted when either:
    \begin{itemize}[nosep, left=0.5em]
        \item Information from either document directly contradicts the hypothesis, or
        \item The combined information from both documents leads to a logical conclusion that contradicts the hypothesis, or
        \item The two documents present mutually contradictory evidence regarding the hypothesis
    \end{itemize}
    
    \item \textbf{Neutral}: The relationship is neutral when:
    \begin{itemize}[nosep, left=0.5em]
        \item Neither document alone nor their combination provides sufficient evidence to support or contradict the hypothesis, or
        \item The documents contain only partially relevant information that doesn't allow for a definitive conclusion, or
        \item The hypothesis introduces new information or claims that go beyond what can be verified from the documents
    \end{itemize}
\end{itemize}

These definitions account for the complex nature of cross-document reasoning, where evidence may be distributed asymmetrically across documents and require different levels of information integration for reaching conclusions.

\paragraph{Hypothesis Creation.}
\label{appendix:hypothesis_generation}

To generate high-quality hypotheses for our CDCL-NLI dataset, we designed a structured prompt for GPT-4o that specified detailed requirements for each label. The complete prompt template is reproduced in Figure \ref{fig:Hypothesises generation}. This prompt design requires GPT-4o to generate evidence explaining the reasoning behind each hypothesis, which significantly reduces hallucination and improves alignment with the source documents. The structured output format facilitates automated processing while ensuring that each hypothesis is accompanied by clear justification of its entailment category. The generated hypotheses were subsequently reviewed by human annotators to ensure quality and adherence to the specified criteria.

\subsection{Data Quality Assessment}
\paragraph{Explanation Annotation Guidelines.}
\label{app:annotation}
We establish the following principles for EDU-based explanation annotation:
\begin{table*}[htbp]
\normalsize
\centering
\begin{tabular}{l|cc}
\toprule
Category & Description (Metric) & Score\\
\midrule
% & Overall IAA (Cohen's $\kappa$) & 0.73 & 0.80 \\
& Entailment (Cohen's $\kappa$) & 0.72 \\
NLI Label & Neutral (Cohen's $\kappa$)& 0.71\\
& Contradiction (Cohen's $\kappa$) & 0.71 \\
\midrule
& EDU Selection (Jaccard Sim.) & 0.76 \\
Explanation & Span Coverage (Overlap Ratio) & 0.81 \\
& Explanation Consistency (Align.) & 0.85 \\
\bottomrule
\end{tabular}
\caption{Dataset quality assessment results.}
\label{tab:annotation_metrics}
\end{table*}
% \begin{table*}[htbp]
% \normalsize
% \centering
% \begin{tabular}{l|ccc}
% \toprule
% Category & Description (Metric) & Init. & Final \\
% \midrule
% % & Overall IAA (Cohen's $\kappa$) & 0.73 & 0.80 \\
% & Entailment (Cohen's $\kappa$) & 0.75 & 0.82 \\
% NLI Label & Neutral (Cohen's $\kappa$) & 0.71 & 0.79 \\
% & Contradiction (Cohen's $\kappa$) & 0.74 & 0.81 \\
% \midrule
% & EDU Selection (Jaccard Sim.) & 0.76 & 0.91 \\
% Explanation & Span Coverage (Overlap Ratio) & 0.81 & 0.94 \\
% & Explanation Consistency (Align.) & 0.85 & 1.00 \\
% \bottomrule
% \end{tabular}
% \caption{Dataset quality assessment results.}
% \label{tab:annotation_metrics}
% \end{table*}

\begin{enumerate}[nosep, leftmargin=*]
    \item \textbf{Minimal Sufficiency}: Annotators should select the minimal set of EDUs that are necessary and sufficient to support the inference conclusion, avoiding redundant or irrelevant units.
    
    \item \textbf{Cross-document Coverage}: Selected EDUs must include evidence from both premise documents when the inference requires cross-document reasoning, ensuring the explanation captures cross-lingual interactions.
    
    \item \textbf{Logical Completeness}: The selected EDUs should form a complete logical chain that clearly demonstrates how the inference conclusion is reached.
\end{enumerate}

\paragraph{Quality Metrics.}
We measured CDCL-NLI dataset using multiple metrics as shown in Table~\ref{tab:annotation_metrics}

The explanation component of our annotations was evaluated using three complementary metrics, all showing exceptional improvement after reconciliation:
\begin{itemize}[nosep, leftmargin=*]
    \item EDU Selection achieved 76\% Jaccard similarity, indicating strong consensus on evidence selection
    \item Span Coverage reached 81\% overlap ratio, demonstrating precise identification of relevant text spans
    \item Explanation Consistency achieved 85\%, ensuring logical coherence in reasoning
\end{itemize}

% \paragraph{Overall Annotation Quality}
Our annotation quality assessment demonstrated strong reliability across all NLI categories. Our initial inter-annotator agreement score is 0.71 and annotation quality is further improved through adjudication.
% Specifically:
% \begin{itemize}[nosep, leftmargin=*]
%     \item Entailment labels achieved the highest final agreement ($\kappa=0.82$)
%     \item Contradiction cases showed strong consensus ($\kappa=0.81$)
%     \item Neutral instances, while slightly lower, maintained robust agreement ($\kappa=0.79$)
% \end{itemize}

Through our rigorous quality control and filtering process, we refined our dataset from an initial collection of 27,750 potential instances to 25,410 high-quality inference pairs. This 8.4\% reduction reflects our commitment to maintaining high standards in both label accuracy and explanation quality, ensuring the dataset's reliability for both classification and interpretability research.

\subsection{Data Information}
\label{app:data_info}

\paragraph{Language Distribution.}

\begin{figure}[htbp]
    \centering
    \includegraphics[width=\linewidth]{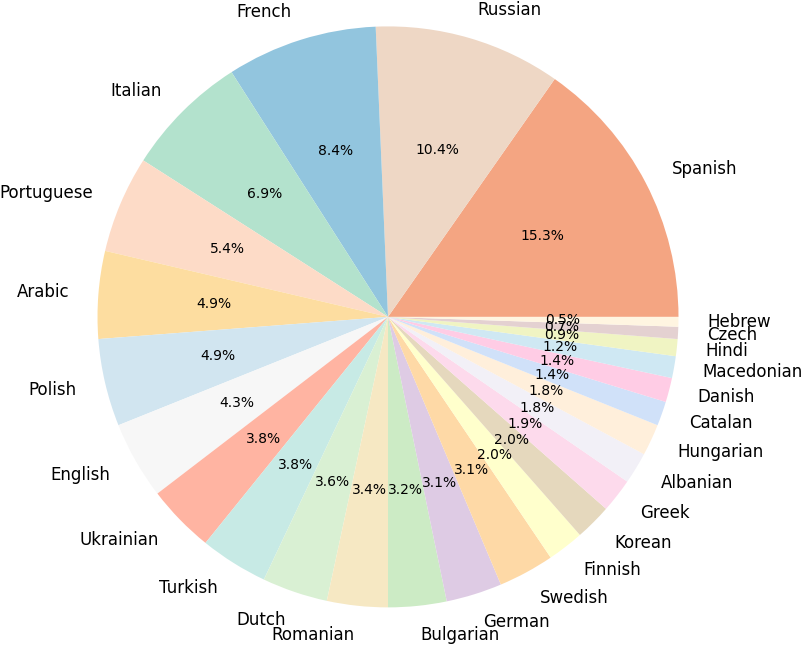}
    \caption{Language distribution of CDCL-NLI dataset.}
    \label{fig:language_distribution}
\end{figure}

Figure~\ref{fig:language_distribution} illustrates the language distribution of our dataset, where Spanish (15.3\%), Russian (10.4\%), and French (8.4\%) represent the top three most frequent languages, while languages like Hebrew, Czech, and Hindi each accounts for approximately 1-2\% of the data. This distribution not only reflects the imbalanced nature of multilingual usage in real-world scenarios but also ensures broad coverage of linguistic phenomena, enabling the study of diverse cross-lingual inference patterns.

\begin{figure*}[htbp]
    \centering
    \begin{subfigure}[t]{0.48\textwidth}
        \centering
        \includegraphics[width=\textwidth]{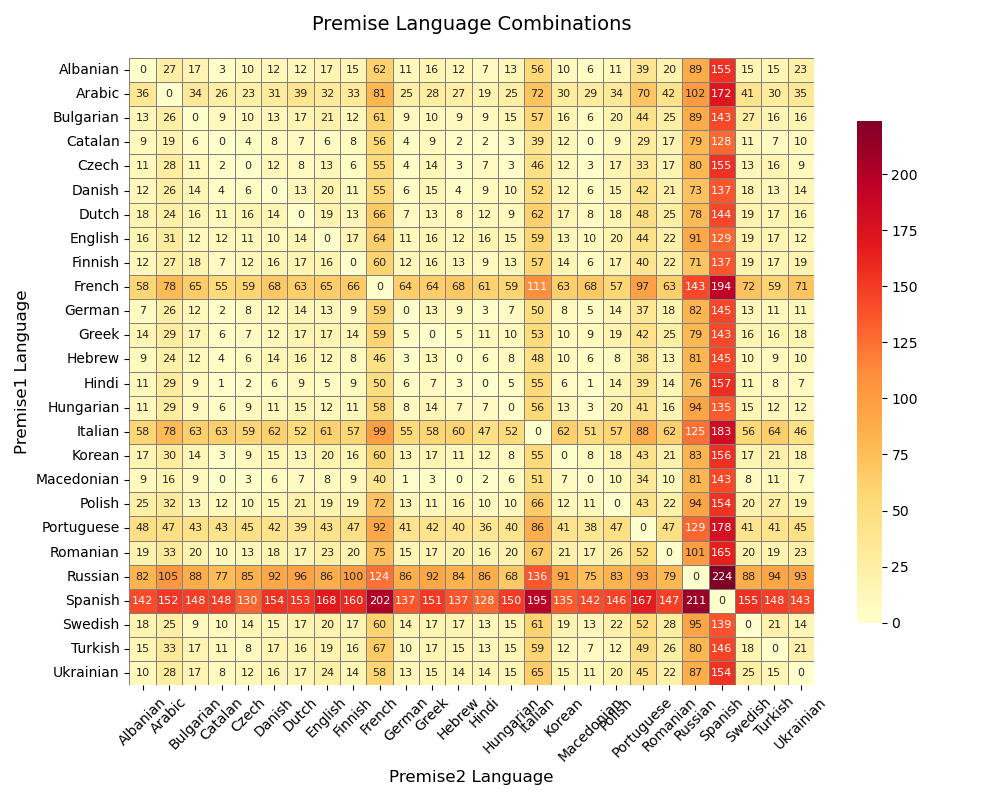}
        \caption{Heat map of premise language combinations across the dataset.}
        \label{fig:lang_heatmap}
    \end{subfigure}
    \hfill
    \begin{subfigure}[t]{0.48\textwidth}
        \centering
        \includegraphics[width=\textwidth]{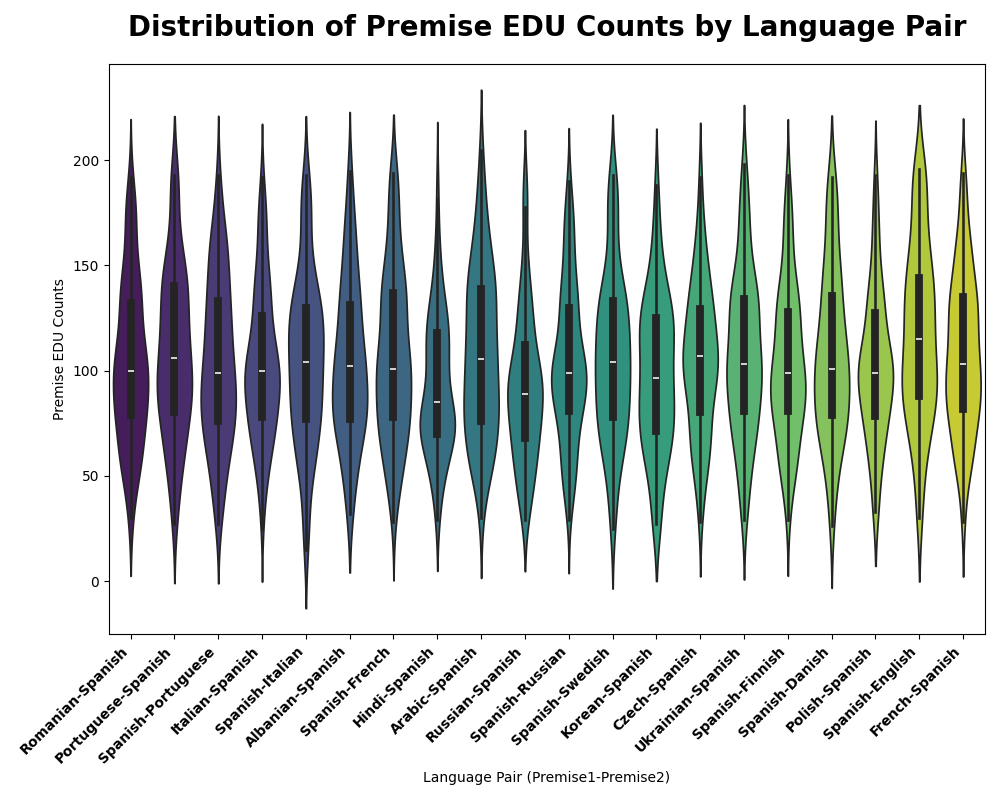}
        \caption{Distributions of EDU counts across top-20 language pairs.}
        \label{fig:edu_violin}
    \end{subfigure}

    \caption{Statistic visualization of language pair distributions and their EDU characteristics.}
    \label{fig:lang_analysis}
\end{figure*}

\paragraph{Language Pair Distribution.}
As shown in Figure~\ref{fig:lang_heatmap}, the dataset exhibits diverse language combinations across 24 languages. Spanish demonstrates the highest interaction frequency with other languages, particularly evident in Spanish-Russian (224 instances) and Spanish-Portuguese (178 instances) pairs. The heat map reveals several interesting patterns:

\begin{itemize}[nosep, leftmargin=*]
    \item Most language pairs maintain a balanced bidirectional relationship, with similar instance counts in both directions
    \item Romance languages (Spanish, French, Portuguese, Italian) show stronger interconnections
    \item Less-resourced languages like Albanian and Macedonian have fewer cross-lingual pairs
    \item Russian and Spanish serve as central hub languages, connecting with most other languages in the dataset
\end{itemize}
\paragraph{EDU Count Distribution by Language Pair.}
The violin plot in Figure~\ref{fig:edu_violin} illustrates the distribution of Elementary Discourse Units (EDUs) across the top language pairs. Several key observations emerge:

\begin{itemize}[nosep, leftmargin=*]
    \item Most language pairs show a median EDU count between 80 and 120 units
    \item The distributions are generally symmetric, indicating consistent EDU patterns regardless of the source language
    \item Romance language pairs (Romanian-Spanish, Portuguese-Spanish, Italian-Spanish) exhibit similar EDU distribution patterns
    \item Some pairs, particularly those involving Spanish as one of the languages, show wider distributions, suggesting more diverse discourse structures
    \item The violin shapes indicate that extreme EDU counts (very low or very high) are relatively rare across all language pairs
\end{itemize}

This analysis suggests that while the dataset maintains diverse language coverage, it also preserves consistent discourse complexity across different language combinations.

\section{Graph Construction Details}
\subsection{Relation Types}
\label{app:graph_relation_types}
\paragraph{RST Graph Construction with Selected Relation Types.}
In constructing individual RST graphs for each document, we select a subset of relation types to focus on the most salient discourse and semantic connections. Specifically, we use the following relation types: Temporal, Summary, Condition, Contrast, Cause, Background, Elaboration, Explanation, and lexical chains. This selection balances coverage and complexity, ensuring that the resulting graph captures essential discourse relations and key semantic links without introducing excessive sparsity or noise. The inclusion of lexical chains further strengthens semantic cohesion by linking related words and expressions across different segments.

\paragraph{Graph Fusion with Extended Relation Types.}
During the fusion of RST graphs from multiple documents, we expand the set of relation types to include a broader range of discourse and organizational structures. The extended set comprises: Temporal, TextualOrganization, Joint, Topic-Comment, Comparison, Condition, Contrast, Evaluation, Topic-Change, Summary, Manner-Means, Attribution, Cause, Background, Enablement, Explanation, Same-Unit, Elaboration, and Lexical chains. This comprehensive set allows for richer cross-document alignment by capturing diverse forms of rhetorical and semantic relationships. Both in single-document and fused graphs, these relations serve as edge types in the construction of the Relation-aware Graph Attention Network (RGAT), enabling the model to effectively encode complex discourse and semantic structures.

\subsection{Node Feature Definition}
\label{app:graph node feature}
Specifically, for leaf nodes, we define:
\[
\phi(v_i) = \phi(\text{EDU}_s), \text{Text}_{v_i} = \text{EDU}_s, \text{Type}_{v_i} = 1.
\]

For branch nodes, we define:
\[
\phi(v_i) = \frac{1}{2}(\phi(v_j) + \phi(v_k)), 
\]
\vspace{-4mm}
\[
 \text{Text}_{v_i} = \text{Text}_{v_j} \oplus \text{Text}_{v_k}, \text{Type}_{v_i} = 0,
\]
where $v_j, v_k$ are the children of $v_i$, and $\oplus$ denotes concatenation.
For completeness, we provide the detailed formulas for the relation-level and node-level attention mechanisms used in updating node embeddings.
\subsection{Justification of the Cross-Document Edge Threshold \(\delta\)}
\label{app:threshold}

\begin{table*}[t]
\centering
\begin{tabular}{lcccccc}
\toprule
Baseline & Base Model & Optimizer & LR & Batch Size & Max Length & Epochs \\
\midrule
Hypothesis-only & XLM-R Large & AdamW & $3 \times 10^{-6}$ & 16 & 512 & 20 \\
DocNLI         & XLM-R Large & AdamW & $3 \times 10^{-6}$ & 16 & 512 & 20 \\
R2F            & XLM-R Large & AdamW & $1 \times 10^{-6}$ & 16 & 512 & 20 \\
Ours            & XLM-R Large & AdamW & $1 \times 10^{-5}$ & 16 & 512(per EDU) & 20 \\
\bottomrule
\end{tabular}
\caption{Training hyperparameters for conventional baseline models and our model. These configurations, including the consistent use of the XLM-RoBERTa-Large base model and AdamW optimizer, were utilized to ensure reproducibility and fair comparison.}
\label{tab:hyperparameters_conventional}
\end{table*}

\begin{figure}[t]
  \centering
  \includegraphics[width=\columnwidth]{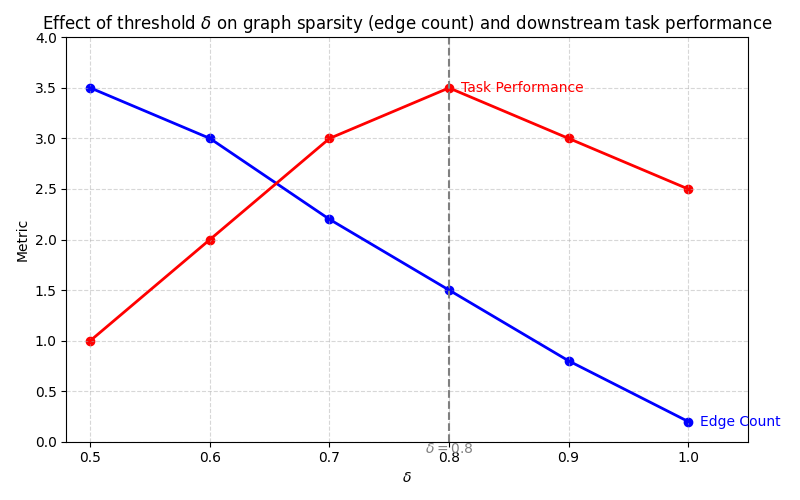}

  \caption{Effect of threshold $\delta$ on graph sparsity and task performance. Edge count (blue) decreases as $\delta$ increases, while task performance (red) peaks at $\delta=0.8$ (dashed line), providing optimal balance between relevant connections and noise reduction.}
  \label{fig:threshold_analysis}

\end{figure}

The threshold \(\delta\) for adding cross-document lexical edges is set to 0.8 based on empirical analysis balancing sparsity and relevance of edges. We evaluated different threshold values on a validation set using the following metrics:

\begin{itemize}[nosep, leftmargin=*]
  \item \textbf{Edge Sparsity}: Higher thresholds reduce the number of edges, leading to sparser graphs that help avoid noise.
  \item \textbf{Semantic Relevance}: Lower thresholds introduce more edges but may include irrelevant or weakly related node pairs.
  \item \textbf{Downstream Task Performance}: We observed that \(\delta=0.8\) achieves the best trade-off, maximizing performance on the target task (e.g., accuracy or F1 score).
\end{itemize}
Figure~\ref{fig:threshold_analysis} shows the impact of varying \(\delta\) on edge count and task performance, confirming the choice of 0.8 as a reasonable and effective threshold.

\subsection{Graph Attention Formulas}
\label{app:attention}
\paragraph{Relation Weight.} The relation importance weights $\alpha_r$ are learnable parameters normalized by softmax:
\[
\alpha_r = \frac{\exp(w_r)}{\sum_{r' \in R} \exp(w_{r'})},
\]
where $w_r$ is a trainable scalar parameter for rhetorical relation $r$.
\paragraph{Hyperparameters.}
For the model defined in Equation~\ref{eq:graph_attention}, the following settings are used:
The first layer uses \(K=4\) attention heads.
The second layer uses \(K=1\) attention head.
Residual connections and dropout with rate 0.1 are applied after each layer.
\paragraph{Node-level Attention Coefficients.} The attention coefficients $\beta_{ij,k}^{r,(l)}$ measure the importance of neighbor node $v_j$ to node $v_i$ under relation $r$, head $k$, and layer $l$. They are computed as:

\begin{equation}
\small
\beta_{ij,k}^{r,(l)} = 
\frac{
\displaystyle \exp\!\left(\psi\!\left(a_{r,k}^{(l)\top}\left[\mathbf{W}_{r,k}\mathbf{h}_{v_i}^{(l-1)}\!\| \mathbf{W}_{r,k}\!\mathbf{h}_{v_j}^{(l-1)}\right]\right)\right)
}{
\displaystyle \sum\limits_{\!v_m \!\in \mathcal{N}_r(v_i)\!} \!\!\!\!\exp\!\left(\psi\!\left(a_{r,k}^{(l)\top\!}\left[\mathbf{W}_{r,k}\mathbf{h}_{v_i}^{(l-1)}\!\| \mathbf{W}_{r,k}\mathbf{h}_{v_m}^{(l-1)}\right]\right)\right)
},
\end{equation}
where $\mathbf{W}_{r,k}$ is the trainable linear transformation matrix for relation $r$ and head $k$, $a_{r,k}^{(l)}$ is the learnable attention vector for relation $r$, head $k$, and layer $l$, $[\cdot \| \cdot]$ denotes vector concatenation, $\psi(\cdot)$ is the ELU activation function.

\paragraph{Additional Details.} Each layer uses residual connections and dropout with a rate of 0.1 to improve training stability. The first layer uses $K=4$ attention heads, while the second layer uses $K=1$.
% This completes the detailed description of the attention mechanism used in our model.

\section{Training Details}
\label{app:training hyperparameters}
\subsection{Model Training Hyperparameters}
\label{app:conventional hyperparameters}
All the models are implemented in PyTorch and trained on an NVIDIA A100 GPU. To ensure fair comparison and reproducibility of results, all conventional baseline models and our model were fine-tuned under consistent experimental settings. As detailed in Table~\ref{tab:hyperparameters_conventional}, each baseline utilizes the \textbf{XLM-RoBERTa-large} pretrained model as the base architecture and the \textbf{AdamW} optimizer for training. The learning rates are carefully selected for each model variant to optimize performance, while maintaining a uniform batch size of 16, a maximum input sequence length of 512 tokens, and training for 20 epochs. These standardized hyperparameters guarantee that performance differences stem from model design rather than training discrepancies, thereby supporting the validity and reproducibility of our comparative evaluation. Specially, for our model, as we split the documents into EDUs, so the maximux length is for one single EDU. By processing shorter EDUs instead of full documents, our model in long-text scenarios minimizes information loss, leading to improved performance.

\subsection{LLM Fine-tuning Hyperparameters}
\label{app:LLM Details}
For fine-tuning the Llama3-8B-instruct and Qwen3-8B model, we employed LoRA (Low-Rank Adaptation) to efficiently adapt the large-scale pretrained model with limited computational resources. The key hyperparameters for LoRA tuning included a rank of 16, which balances adaptation capacity and parameter efficiency, and a dropout rate of 0.1 to mitigate overfitting. The learning rate was set to $2 \times 10^{-4}$ with a linear warmup over the first 500 steps, followed by a constant decay. We used a batch size of 64 sequences and capped the maximum input length at 1024 tokens to fully leverage the model’s context window. Training was conducted for 10 epochs, which empirically provided a good trade-off between convergence and training cost. These hyperparameters were chosen based on prior LoRA tuning best practices and preliminary experiments to ensure stable and effective adaptation of the Llama3-8B-instruct and Qwen3-8B model. The prompt is shown in Figure \ref{fig:ft prompt}.

\begin{figure*}[t]
\begin{response}{Fine-tunning Prompt}

You are skilled in the NLI task. Given a premise consisting of two documents and a hypothesis, each with its specified language, your task is to determine the natural language inference (NLI) relationship between the hypothesis and the premise. Note that the premise and hypothesis may be in different languages. The output should be one of three labels: \texttt{Entailment}, \texttt{Contradiction}, or \texttt{Neutral}.

Input format:

Premise 1 (Language: \textless Lang1\textgreater): \textless Premise1 text\textgreater

Premise 2 (Language: \textless Lang2\textgreater): \textless Premise2 text\textgreater

Hypothesis: \textless Hypothesis text\textgreater

Output format:

One of the labels: \texttt{Entailment}, \texttt{Contradiction}, or \texttt{Neutral}

---

Example:

Premise 1 (Language: English): The cat is sitting on the mat.

Premise 2 (Language: French): Le chat est assis sur le tapis.

Hypothesis: The animal is resting on a rug.

Output: Entailment

---

Now, given the input premises and hypothesis, provide the NLI label.
\end{response}

\caption{Llama3-8B-Instruct and Qwen3-8B Finetuning Prompt.}
\label{fig:ft prompt}

\end{figure*}

\begin{table*}[h]
\centering
\begin{tabular}{l|ccc|ccc}
\toprule
\textbf{Model} & \multicolumn{3}{c|}{\textbf{TestSet1: Cross-Lingual}} & \multicolumn{3}{c}{\textbf{TestSet2: English}} \\
               & Precision & Recall & F1 Macro & Precision & Recall & F1 Macro \\
\midrule
Llama-3-8B     & 44.00     & 50.00  & 46.00    & 49.00     & 55.00  & 50.00    \\
GPT-4o         & 50.00     & 54.00  & 52.00    & 59.00     & 62.00  & 61.00    \\
Qwen3-8B       & 58.00     & 54.00  & 57.00    & 68.00     & 64.00  & 63.00    \\
\bottomrule
\end{tabular}

\caption{Zero-shot performance of large language models on the CDCL-NLI dataset.}
\label{tab:zeroshot_results}

\end{table*}

\begin{table*}[h]
\centering
\begin{tabular}{lccc}
\toprule
\textbf{Model} & \textbf{Single Document1} & \textbf{Single Document2} & \textbf{Combined Documents} \\
\midrule
DocNLI & 54.22 & 54.95 & 64.46 \\
R2F    & 57.09 & 57.12 & 65.42 \\
\bottomrule
\end{tabular}

\caption{F1 Macro scores for different methods across premises with varying numbers of documents.}
\label{tab:single_doc_f1_basaelins}
\end{table*}

\begin{table*}[t]
\centering
\small
\begin{tabular}{p{0.05\textwidth} p{0.4\textwidth} | p{0.05\textwidth} p{0.38\textwidth}}
\toprule
\textbf{EDU} & \textbf{Text} & \textbf{EDU} & \textbf{Text} \\
\midrule
1  & 7. května & 22 & řekl prokurátor Giovanni Matos místní televizní stanici Canal N. \\
4  & Společnost okamžitě nereagovala na žádost o komentář. & 24 & jsou 27 obětí,“ \\
7  & (Reuters) - & 25 & „Informace jsou správné, \\
$11^{\textcolor{red}{\textcircled{\scriptsize 1}}}$ & \textcolor{darkgreen}{Úřadníci uvedli v neděli, že nehoda v malé zlaté dolině na jihu Peru odnesla život 27 pracovníků.} & 26 & potvrdila je policie v Yanaquihuě, \\
12 & Jedná se o jeden z nejúmrtnějších důležitých událostí v těžebním průmyslu v tomto jihoamerickém státě. & 27 & „Jedná se o formální dolinu (...), \\
{$15^{\textcolor{red}{\textcircled{\scriptsize 2}}}$} & \textcolor{darkgreen}{Nehoda se stala v sobotu ráno v těžební společnosti Yanaquihua, která se nachází v provincii Condesuyos v departementu Arequipa.} & 30 & dodal. \\
17 & \textcolor{darkgreen}{Zdá se, že došlo ke zkratu, která způsobila požár uvnitř tunelu,} & 33 & musíme jít \\
18 & uvedla regionální vláda. & 34 & a zjistit, kde jsou mrtví, jestli je tam bezpečné, \\
{$37^{\textcolor{red}{\textcircled{\scriptsize 3}}}$} & \textcolor{darkgreen}{Regionální vláda Arequipy a ministerstvo vnitra mobilizovaly policie, zdravotníky a sanitky, aby pomohly při péči o oběti a jejich záchraně. }& 35 & aby se tam mohli dostat policisté a soudní pracovníci \\
39 & Podle statistik peruánského ministerstva těžeb a energie je toto nejvyšší počet obětí v jediném těžebním nehodě & 36 & a provést procedury,“ \\
40 & nejméně od roku 2000. & & \\
\bottomrule
\end{tabular}
\caption{Elementary Discourse Units (EDUs) from $Document_1$ with their corresponding Spanish text. Segments highlighted in \textcolor{darkgreen}{green} represent evidence supporting the Entailment classification. EDU indexes with circled numbers {\textcolor{red}{\textcircled{\scriptsize 1}}} indicate cross-document "Lexical" chains linking to corresponding EDUs in $Document_2$.}
\label{tab:doc1_edu}

\end{table*}

\section{Additional Experiments}
\subsection{LLM Few-shot Prompt}
\label{app:evaluation_prompt}
%The validation prompt is designed for the cross-document and cross-language Natural Language Inference (NLI) task, where the input consists of two news articles in different languages serving as premises, along with a hypothesis sentence. 
As shown in Figure \ref{fig:validation_prompt}, one example is provided to demonstrate how to determine the logical relationship between the premise and the hypothesis. The model is instructed to output exactly one of three labels: \textit{entailment}, \textit{contradiction}, or \textit{neutral}. This prompt effectively guides the model to understand the task objective and output format, thereby enhancing its reasoning capability across multiple languages and documents during the few-shot validation stage \cite{10.1162/tacl_a_00697}.

\subsection{LLM in Zero-shot Scenario}
\label{app: LLM_zero}

The zero-shot results reported in Table~\ref{tab:zeroshot_results} are obtained using the same prompt design as the few-shot experiments, differing only in the absence of in-context examples. As expected, all models perform worse under the zero-shot setting compared to their few-shot counterparts, demonstrating the effectiveness and necessity of providing exemplars in the prompt for this task. Despite the overall performance drop, the relative ranking of the three models remains consistent with the few-shot scenario, with Qwen3-8B achieving the highest scores, followed by GPT-4o, and then Llama-3-8B. This consistency indicates that these models’ capabilities in handling the CDCL-NLI task are stable across different prompting strategies. Moreover, the results highlight the challenge of zero-shot cross-document and cross-lingual natural language inference, emphasizing the importance of prompt engineering and in-context learning to boost model performance on complex multilingual and multi-document reasoning tasks.

\subsection{Baseline Evaluation in Single Document Scenario}
\label{app: baseline_singledoc}

To further demonstrate the cross-document characteristic of our dataset, we add this extra experiment to evaluate the performance using either a single document ($Document_1$ or $Document_2$) as the premise compared to using the full combined premise, as summarized in Table~\ref{tab:single_doc_f1_basaelins}. The noticeable improvement in F1 score when both documents are combined indicates that effective inference relies on integrating information from multiple sources.
Additionally, the similar results observed between $\textbf{Single Document 1}$ (54.22\% and 57.09\% F1) and $\textbf{Single Document 2}$ (54.95\% and 57.12\% F1) imply that each document provides valuable and roughly equal contributions. This further supports the notion that reasoning in this task benefits from synthesizing evidence across documents rather than focusing on a single source.

\section{Case Study}
\label{app:case study}

\subsection{Our Method Case}

\begin{table*}[h]
\centering
\small
\begin{tabular}{p{0.05\textwidth} p{0.4\textwidth} | p{0.05\textwidth} p{0.4\textwidth}}
\toprule
\textbf{EDU} & \textbf{Text} & \textbf{EDU} & \textbf{Text} \\
\midrule
14 & informó el Ministerio Público de ese país. & 53 & [Al menos siete muertos en Texas \\
{$15^{\textcolor{red}{\textcircled{\scriptsize 1}}}$} & \textcolor{darkgreen}{Al menos 27 personas murieron en Perú} & 54 & tras atropellamiento en una parada de autobús cerca de un refugio para inmigrantes] \\
17 & y otras dos fueron rescatadas & 56 & lo que impidió que los mineros pudieran escapar. \\
18 & \textcolor{darkgreen}{luego de un incendio el sábado en una mina de oro en la sureña provincia de Condesuyos,} & 57 & Se informó que \\
21 & Según las primeras investigaciones, la tragedia tuvo lugar & 59 & el fuego se propagó de manera muy rápida por las estructuras de madera que sostienen el yacimiento, \\
{$23^{\textcolor{red}{\textcircled{\scriptsize 2}}}$} & \textcolor{darkgreen}{tras producirse un cortocircuito a 100 metros de la entrada de la mina Yanaquihua,} & 60 & dedicado a la extracción de oro, \\
24 & conocida como Esperanza I. & 61 & Medios locales peruanos indicaron que \\
28 & informó el Gobierno regional de Arequipa. & 63 & 27 trabajadores quedaron atrapados en la mina \\
29 & \textcolor{darkgreen}{“Se habría producido un cortocircuito} & 64 & tras un incendio. \\
31 & que provocó un incendio en el interior del socavón, & 65 & Getty Images \\
32 & que habría puesto en riesgo la vida de los trabajadores”, & 71 & James Casquino, alcalde de Yanaquihua, dijo que \\
33 & Medios locales indicaron que & 73 & el dueño de la mina fue a la comisaría de ese distrito \\
34 & \textcolor{darkgreen}{27 trabajadores atrapados habían fallecido por asfixia.} & 75 & para pedir ayuda en el rescate de las personas \\
35 & La noche del sábado, el Ministerio del Interior confirmó en su cuenta de Twitter el accidente. & 76 & que se encontraban atrapadas. \\
38 & indicó el tuit. & 78 & [Mueren varios migrantes en un accidente de auto en Nuevo México cerca de la frontera] \\
39 & “Personal policial se encuentra en el distrito de Yanaquihua & 79 & Las autoridades indicaron que \\
41 & para apoyar en las labores de rescate de los cuerpos de mineros & {$80^{\textcolor{red}{\textcircled{\scriptsize 3}}}$} & \textcolor{darkgreen}{hacia la zona se habían movilizado rescatistas.} \\
42 & que fallecieron dentro de un socavón en la provincia de Condesuyos”, & 81 & Familiares de las víctimas se reunieron frente a la comisaría de Yanaquihua \\
49 & Imágenes difundidas en redes sociales mostraban una gran columna de humo negro proveniente de la mina, & 83 & para recabar información sobre la suerte de sus seres queridos \\
51 & y medios locales indicaron que & 84 & y exigir a las autoridades que agilizaran las labores de rescate de los cuerpos. \\
52 & en el momento del cortocircuito había personal trabajando a unos 80 metros de profundidad. & 85 & El fiscal Giovanni Matos indicó a un medio local que \\
87 & las tareas en la mina podían demorar & 89 & porque no se sabía si los equipos de rescatistas podían ingresar a la mina \\
23 & para retirar los cadáveres. & 90 & para retirar los cadáveres. \\
91 & [Una tormenta de polvo en Illinois causa múltiples muertes y decenas de hospitalizados tras choque masivo] & 94 & indica la compañía en su página web. \\
95 & La mina pertenece a Yanaquihua S. A. C., una empresa & 96 & que reúne a pequeños productores mineros dedicados a la explotación del oro y otros metales, \\
\bottomrule
\end{tabular}

\caption{Elementary Discourse Units (EDUs) from $Document_2$ with their corresponding Spanish text. Segments highlighted in \textcolor{darkgreen}{green} represent evidence supporting the Entailment classification. EDU indexes with circled numbers {\textcolor{red}{\textcircled{\scriptsize 1}}} indicate cross-document "Lexical" chains linking to corresponding EDUs in $Document_1$.}

\label{tab:doc2_edu}
\end{table*}

%\begin{table}[h]
%\centering
%\small
%\begin{tabular}{cl}
%\toprule
%\textbf{Doc} & \textbf{EDU IDs} \\
%\midrule
%1 & 11, 15, 17, 37 \\
%2 & 15, 18, 23, 29, 34, 80 \\
%\bottomrule
%\end{tabular}
%\caption{Extracted EDU IDs used as explanations by the model.}
%\label{tab:explanation_edu}
%\end{table}
Our approach employs a multi-stage framework for analyzing complex multi-document multi-lingual NLI scenarios. Take the given example in Figure \ref{fig:validation_prompt}, the Yanaquihua gold mine incident in Condesuyos, Peru, where a short circuit-induced fire resulted in 27 fatalities among workers trapped within a tunnel, prompting mobilization of local authorities and rescue teams. We begin by parsing the premise documents using Rhetorical Structure Theory (RST), which generates hierarchical discourse trees wherein each node represents an Elementary Discourse Unit (EDU). These nodes are assigned unique indices, with their textual content comprehensively documented in Tables~\ref{tab:doc1_edu} and~\ref{tab:doc2_edu}.

Following RST parsing, we construct individual discourse graphs for each premise document. These discrete graphs are subsequently integrated into a unified premise graph through the establishment of "Lexical" chains that leverage semantic information and discourse relations to facilitate enhanced inference. As illustrated in Tables~\ref{tab:doc1_edu} and~\ref{tab:doc2_edu}, EDU nodes sharing identical uppercase character designations indicate the presence of cross-document ``Lexical'' chains. This consolidated graph representation effectively captures the comprehensive discourse context across the premises, enabling more robust and coherent semantic modeling.

The classification module processes this unified graph in conjunction with the hypothesis to predict the appropriate NLI label. Concurrently, the explanation extraction module identifies a salient subset of nodes within the premise graph that substantiate the classification decision. These explanation nodes are visually distinguished through green font highlighting in Tables~\ref{tab:doc1_edu} and~\ref{tab:doc2_edu}, explicitly denoting their explanatory significance.

Our integrated methodology capitalizes on the hierarchical discourse structure inherent in RST parsing and the semantic connectivity across documents, ensuring that the model's inference is both accurate and interpretable. The explicit identification of explanation nodes within the discourse structure facilitates transparent, human-comprehensible rationales grounded in the premise texts, thereby advancing the explainability of NLI systems in complex multi-document, multi-lingual scenarios. This approach proves particularly valuable when analyzing intricate real-world situations such as the Yanaquihua mine disaster, where understanding the causal relationships and contextual factors is crucial for proper inference.

\subsection{LLM Answer Case}
As shown in Table~\ref{tab:main_results}, Qwen3-8B achieves higher scores compared to Llama3-8B-instruct and the closed-source GPT-4o. One key reason is that we evaluate Qwen3-8B using its \texttt{thinking} (chain-of-thought) mode, as illustrated in Figure~\ref{fig:qwen3-cot}. We still take the case in validation prompt(Tabel \ref{fig:validation_prompt}) as an example, the model systematically parses each premise, accurately extracts key facts, and performs detailed cross-checking between the articles and the hypothesis. It also demonstrates the ability to handle subtle differences in wording (such as distinguishing between deaths and rescues) and to resolve potential ambiguities in translation (e.g., the meaning of "obětí" in Czech). 

Nevertheless, our proposed approach still outperforms Qwen3-8B, primarily due to its ability to explicitly capture document structure through RST parsing and cross-document, cross-lingual semantic integration via "Lexical" chains. Moreover, our method demonstrates superior efficiency with significantly lower computational requirements and faster inference time, making it more practical for real-world applications while maintaining state-of-the-art performance.

\begin{figure*}[ht]
\begin{response}{Hypothesis Generation Prompt}
\textbf{[Hypothesis Generation Prompt]}
We are creating a cross-document cross-lingual NLI dataset. Below are two documents under the event topic: [CATEGORY], treated as one premise in this NLI task. Based on them, generate hypotheses in three labels.
You must \textbf{strictly follow} the instructions:

\textbf{1. Hypothesis:}
The hypothesis should be a factual statement based on the content of the articles. It must be a simple statement and \textbf{should not contain any explanation or analysis like ``this contradicts'' or ``this agrees with'' or ``this is inconsistent with.''}

\textbf{2. Evidence:}
The evidence section should explain how the hypothesis relates to the articles, including any contradictions or confirmations, using specific quotes from the articles.

\textbf{Document Details:}
\begin{itemize}[nosep, leftmargin=*]
    \item \textbf{Document 1}: Date: [DATE\_1]; Article: [ARTICLE\_1]
    \item \textbf{Document 2}: Date: [DATE\_2]; Article: [ARTICLE\_2]
\end{itemize}

\textbf{[Task 1: Entailment Generation]
Generate an Entailment Hypothesis and evidence.}

The hypothesis is supported if evidence from both documents together or from one document alone (without contradiction in the other) logically supports it.  

\textbf{Guidelines:}
\begin{itemize}[nosep, leftmargin=*]
    \item Ensure each detail is verifiable by premise
    \item Include specific facts (dates, names, etc.)
    \item No speculation—strictly based on facts
\end{itemize}

\textbf{Evidence:}
\begin{itemize}[nosep, leftmargin=*]
    \item Quote relevant parts from both articles and explain how they jointly support the hypothesis
\end{itemize}

\textbf{[Task 2: Neutral Generation]}
Generate a Neutral Hypothesis and evidence.

One hypothesis is neutral if there is insufficient or only partial evidence in the premise to confirm or deny it, or if it contains information beyond what the premise verify.

\textbf{Guidelines:}
\begin{itemize}[nosep, leftmargin=*]
    \item Reasonable speculation or expanded related aspects in a reasonable way
    \item Propose middle ground if there's conflicting information
\end{itemize}

\textbf{Evidence:}
\begin{itemize}[nosep, leftmargin=*]
    \item Show partial support from one or both articles without full confirmation
    \item Explain how the hypothesis goes beyond but stays consistent with the Document content
\end{itemize}

Remember, A neutral hypothesis should not be directly confirmed by the premise (which would make it entailed), nor should it contradict the articles (which would make it conflicting).

\textbf{[Task 3: Conflicting Generation]}
Generate a Conflicting Hypothesis and evidence.

One hypothesis is contradicted if either document or their combined information directly opposes it, or if the documents conflict with each other regarding the hypothesis.  

\textbf{Guidelines:}
\begin{itemize}[nosep, leftmargin=*]
    \item Negate or reverse key information in premise
    \item Complex and multi-faceted hypothesis with multiple contradictions
    \item Try to combine multiple points of contradiction
    \item Ensure the hypothesis appears reasonable but actually conflicts clearly
\end{itemize}

\textbf{Evidence:}
\begin{itemize}[nosep, leftmargin=*]
    \item Show which document(s) the hypothesis contradicts and explain specific points
    \item If applicable, explain why this hypothesis cannot coexist with the premise content
\end{itemize}

Output in JSON format:
\begin{mdframed}[backgroundcolor=gray!10, roundcorner=2pt, linewidth=0.5pt, innertopmargin=2pt, innerbottommargin=2pt, innerleftmargin=2pt, innerrightmargin=2pt]
\footnotesize
\begin{verbatim}
{ "entail_evidence": "...",
  "entail_hypothesis": "...",
  "neutral_evidence": "...",
  "neutral_hypothesis": "...",
  "conflict_evidence": "...",
  "conflict_hypothesis": "..."}
\end{verbatim}
\end{mdframed}
\end{response}
\vspace{-3mm}
\caption{Hypothesises Generation Prompt.}
\label{fig:Hypothesises generation}
\end{figure*}

\begin{figure*}[t]
% \begin{response}
\begin{promptbox}[Validation Prompt]
You are tasked with a cross-document and cross-language Natural Language Inference (NLI) task. Your goal is to determine the relationship between the "premise" and the "hypothesis". The premise consists of two documents presented in different languages.
Here is one example:

\begin{tcolorbox}[title=Premise(Document1 in Spanish)]
% Al menos 27 personas murieron en Perú y otras dos fueron rescatadas luego de un incendio el sábado en una mina de oro en la sureña provincia de Condesuyos, informó el Ministerio Público de ese país.
% Según las primeras investigaciones, la tragedia tuvo lugar tras producirse un cortocircuito a 100 metros de la entrada de la mina Yanaquihua, conocida como Esperanza I.
% “Se habría producido un cortocircuito que provocó un incendio en el interior del socavón, que habría puesto en riesgo la vida de los trabajadores”, informó el Gobierno regional de Arequipa.
% Medios locales indicaron que 27 trabajadores atrapados habían fallecido por asfixia.
% La noche del sábado, el Ministerio del Interior confirmó en su cuenta de Twitter el accidente.
% Personal policial se encuentra en el distrito de Yanaquihua para apoyar en las labores de rescate de los cuerpos de mineros que fallecieron dentro de un socavón en la provincia de Condesuyos.
% Imágenes difundidas en redes sociales mostraban una gran columna de humo negro proveniente de la mina, y medios locales indicaron que en el momento del cortocircuito había personal trabajando a unos 80 metros de profundidad.
% Se informó que el fuego se propagó de manera muy rápida por las estructuras de madera que sostienen el yacimiento, dedicado a la extracción de oro, lo que impidió que los mineros pudieran escapar.James Casquino, alcalde de Yanaquihua, dijo que el dueño de la mina fue a la comisaría de ese distrito para pedir ayuda en el rescate de las personas que se encontraban atrapadas.

Al menos 27 personas murieron y dos fueron rescatadas tras un incendio en la mina de oro Yanaquihua, en Condesuyos, Perú. Las investigaciones apuntan a un cortocircuito ocurrido a unos 100 metros de la entrada, que provocó un fuego que se expandió rápidamente por las estructuras de madera del socavón, dificultando la salida de los trabajadores. Autoridades regionales señalaron que la mayoría de los mineros fallecieron por asfixia. El Ministerio del Interior confirmó el accidente la noche del sábado y la policía trabaja en el rescate de los cuerpos. Imágenes difundidas mostraron una densa columna de humo, mientras que medios locales informaron que había personal laborando a 80 metros de profundidad. El alcalde de Yanaquihua indicó que el dueño de la mina acudió a la comisaría para solicitar apoyo en el rescate.

\end{tcolorbox}

\begin{tcolorbox}[title=Premise(Document2 in Czech)]

7. května (Reuters) - Úřadníci uvedli v neděli, že nehoda v malé zlaté dolině na jihu Peru odnesla život 27 pracovníků. Jedná se o jeden z nejúmrtnějších důležitých událostí v těžebním průmyslu v tomto jihoamerickém státě.
Nehoda se stala v sobotu ráno v těžební společnosti Yanaquihua, která se nachází v provincii Condesuyos v departementu Arequipa. Zdá se, že došlo ke zkratu, která způsobila požár uvnitř tunelu, uvedla regionální vláda.
„Informace jsou správné, potvrdila je policie v Yanaquihuě, jsou 27 obětí,“ řekl prokurátor Giovanni Matos místní televizní stanici Canal N.
Regionální vláda Arequipy a ministerstvo vnitra mobilizovaly policii, zdravotníky a sanitky, aby pomohly při péči o oběti a jejich záchraně.
Podle statistik peruánského ministerstva těžeb a energie je toto nejvyšší počet obětí v jediném těžebním nehodě nejméně od roku 2000.
Společnost okamžitě nereagovala na žádost o komentář.
\end{tcolorbox}
\begin{tcolorbox}[title=Hypothesis and Label]
Hypothesis: "The Yanaquihua gold mine in the province of Condesuyos, Peru, experienced a fatal incident on a Saturday morning due to a short circuit that caused a fire, leading to the death of 27 workers trapped inside the tunnel, with local authorities and rescue teams being mobilized to the site."

Output: entailment
\end{tcolorbox}

Now, given the following input, output exactly one word indicating the NLI label: \texttt{neutral}, \texttt{entailment}, or \texttt{contradiction}.

Premise:[Premsise]

Hypothesis:
[Hypothesis]

Output the NLI label (neutral/entailment/contradiction):
\end{promptbox}
\vspace{-4mm}
\caption{LLM Validation Prompt.}
\label{fig:validation_prompt}
\end{figure*}

\begin{figure*}[ht]
\centering
\begin{response}{Qwen3-8B Reasoning Output}

Okay, let's tackle this NLI task. So, the premise has two articles in Spanish and Czech, and the hypothesis is in English. I need to determine if the hypothesis is entailed, contradicted, or neutral based on the premises.
First, I'll read through the Spanish article. It mentions that on Saturday, there was a fire in a gold mine in Condesuyos province, caused by a short circuit 100 meters from the entrance. The mine is called Yanaquihua, known as Esperanza I. They say 27 people died, two were rescued. The regional government of Arequipa mentioned the short circuit caused the fire, risking workers' lives. Local media said 27 trapped workers died of asphyxia. The Ministry of Interior confirmed the accident on Saturday, and police were there for rescue. The fire spread quickly through wooden structures, preventing escape. The mine belongs to Yanaquihua S.A.C., a company with small miners.

Now the Czech article: Officials said the accident in the small gold mine in southern Peru resulted in 27 deaths. It happened on Saturday morning at the Yanaquihua mine in Condesuyos province, Arequipa department. A short circuit caused a fire inside the tunnel. The regional government confirmed 27 victims. The prosecutor Giovanni Matos said the information is correct, and police are there to find the dead, check safety for rescue teams. Regional government and interior ministry mobilized police, medics, and ambulances. It's the highest number of deaths in a single mining accident since 2000. The company didn't respond to a comment request.

Hypothesis: The Yanaquihua gold mine in Condesuyos, Peru, had a fatal incident on Saturday morning due to a short circuit causing a fire, leading to 27 workers' deaths trapped inside the tunnel, with local authorities and rescue teams mobilized.
Check each part. The Spanish article says the fire was caused by a short circuit, leading to 27 deaths. The Czech article confirms the same: short circuit causing fire, 27 deaths. Both mention Saturday morning (the Spanish says "sábado" which is Saturday, and the Czech says "v sobotu ráno" which is Saturday morning). The mine is Yanaquihua in Condesuyos, Peru. The death count is 27, and both mention authorities and rescue teams being mobilized. The hypothesis states all these points. The Spanish article also mentions the mine's name and the company, which the Czech article doesn't, but that's not conflicting. The hypothesis doesn't mention the company, so that's okay. 

Are there any contradictions? The Spanish article mentions two people rescued, but the hypothesis says 27 died. Wait, the hypothesis says "leading to the death of 27 workers trapped inside the tunnel." The Spanish article says 27 died and two were rescued. The Czech article says 27 victims, which could include the two rescued? Wait, maybe not. If two were rescued, then the deaths would be 27, and the two rescued are alive. So the hypothesis says 27 workers died, which matches the Spanish article's 27 deaths. The Czech article says "27 obětí" which is victims, which could be the 27 who died, but maybe the two rescued are not counted as victims? Or maybe the Czech article is using "obětí" as victims, which could include those who died and those rescued? Wait, in Czech, "obětí" is victims, usually meaning those who died. But the Spanish article says 27 died and two rescued. The Czech article says 27 victims, which might be the same 27, but the two rescued might not be considered victims. So the hypothesis says 27 workers died, which aligns with both articles. The Czech article says "27 obětí" which is victims, so that's consistent. 

Also, the hypothesis mentions the fire caused by a short circuit, which both articles confirm. The mobilization of authorities and rescue teams is mentioned in both. The time is Saturday morning in both. The mine's name and location are correct. 

Therefore, the hypothesis is fully supported by both articles. So the label should be entailed.
\end{response}
\vspace{-4mm}
\caption{Illustration of Qwen3-8B's chain-of-thought reasoning process. The model systematically analyzes premise documents by extracting key facts, performing cross-document verification, and resolving translation ambiguities (e.g., Czech term "obětí"). While this reasoning approach enables Qwen3-8B to outperform other LLMs like Llama3-8B-instruct and GPT-4o, our RST-based method achieves superior performance through more intuitive explanations, explicit discourse structure modeling, and cross-lingual semantic integration, all while requiring significantly lower computational resources and faster processing times.}

\label{fig:qwen3-cot}
\end{figure*}

\end{document}